%% file: 000_main_icml.tex
\documentclass{article}

\usepackage{xcolor} 

\usepackage{microtype}
\usepackage{graphicx}
\usepackage{booktabs} 

\usepackage{hyperref}

\usepackage{paralist}

\usepackage[accepted]{icml2021}

\icmltitlerunning{Thinking Like Transformers}

\input{tex_preamble}

\begin{document}

\twocolumn[
\icmltitle{Thinking Like Transformers}

\icmlsetsymbol{equal}{*}

\begin{icmlauthorlist}
\icmlauthor{Gail Weiss}{technion}
\icmlauthor{Yoav Goldberg}{barilan,ai2}
\icmlauthor{Eran Yahav}{technion}
\end{icmlauthorlist}

\author{%
	Gail Weiss \\
	Technion\\
	\texttt{sgailw@cs.technion.ac.il} \\
	\And
	Yoav Goldberg \\
	Bar Ilan University \\
	Allen Institute for AI\\
	\texttt{yogo@cs.biu.ac.il} \\
	\And
	Eran Yahav\\
	Technion \\
	\texttt{yahave@cs.technion.ac.il} \\
}

\icmlaffiliation{technion}{Technion, Haifa, Israel}
\icmlaffiliation{barilan}{Bar Ilan University, Ramat Gan, Israel}
\icmlaffiliation{ai2}{Allen Institute for AI}

\icmlcorrespondingauthor{Gail Weiss}{sgailw@cs.technion.ac.il}

\icmlkeywords{Transformers, abstraction, DSL}

\vskip 0.3in
]

\printAffiliationsAndNotice{}  

\input{00_abstract}
\input{01_introduction}
\input{02_overview}

\input{03_rasp}

\input{04_implications}
\input{05_experiments}

\input{06_conclusions}

\smallparagraph{Acknowledgments}
We thank Uri Alon, Omri Gilad, Daniel Filan, and the  reviewers for their constructive comments. This project received funding from European Research Council under European Union's Horizon 2020 research \& innovation programme, agreement $\#$802774 (iEXTRACT).

\bibliography{bib}
\bibliographystyle{icml2021}

\input{App00_all}

\end{document}

%% file: tex_preamble.tex
\usepackage{amsmath} 
\usepackage{cleveref} 
\usepackage{amsthm} 
\usepackage{listings}
\usepackage[most]{tcolorbox}
\usepackage{inconsolata}

\input{math_commands}

\input{genericmacros}
\input{specificmacros}

\usepackage{url} 

\usepackage[utf8]{inputenc} 
\usepackage[T1]{fontenc}    
\usepackage{booktabs}       
\usepackage{amsfonts}       
\usepackage{nicefrac}       
\usepackage{microtype}      
\usepackage{diagbox} 
\usepackage{multirow}

\usepackage[inline]{enumitem} 
\usepackage{framed} 
\usepackage{mathtools}
\usepackage{amssymb} 
\usepackage[inline]{enumitem}
\usepackage{subfig}
\usepackage{relsize} 

\usepackage{stfloats}

%% file: math_commands.tex
\usepackage{amsmath,amsfonts,bm}

\def\eqref#1{equation~\ref{#1}}

\def\1{\bm{1}}

\DeclareMathAlphabet{\mathsfit}{\encodingdefault}{\sfdefault}{m}{sl}
\SetMathAlphabet{\mathsfit}{bold}{\encodingdefault}{\sfdefault}{bx}{n}



%% file: genericmacros.tex
\usepackage{ifthen}

\newcommand{\pcite}[1]{\citeauthor{#1} \yrcite{#1}}

\newcommand{\ignore}[1]{}

\newcommand{\DONE}[1]{}
\newcommand{\revisionhide}[1]{}

\crefname{lemma}{lemma}{lemmas} 
\usepackage{pifont}
\newcommand{\xmark}{\ding{55}}%
\newcommand{\halfcheck}{\checkmark\kern-1.1ex\raisebox{.7ex}{\rotatebox[origin=c]{125}{--}}}

\newcommand\tstrut{\rule{0pt}{2.6ex}}         

\usepackage{newunicodechar}
\newunicodechar{§}{\makeabbreviationtt}
\def\makeabbreviationtt#1§{\texttt{#1}}
\newcommand{\bos}{\S}

\newcommand{\smallparagraph}[1]{{\bf #1\ }}

%% file: specificmacros.tex
\newcommand{\GW}[1]{ {\color{blue} \bf (GW: #1)} }
\newcommand{\EY}[1]{ {\color{purple} \bf (EY: #1)} }
\newcommand{\YG}[1]{ {\color{brown} \bf (YG: #1)} }

\renewcommand{\GW}[1]{ }
\renewcommand{\EY}[1]{ }
\renewcommand{\YG}[1]{ }

\newcommand{\rasp}{{\small \textsc{RASP}}}
\newcommand{\encoder}{s-op}
\newcommand{\encoders}{s-ops}
\newcommand{\encoderfull}{sequence operators}

%% file: 00_abstract.tex
\begin{abstract}
What is the computational model behind a Transformer?
Where recurrent neural networks have direct parallels in finite state machines, allowing clear discussion and thought around architecture variants or trained models, Transformers have no such familiar parallel.
In this paper we aim to change that, proposing a computational model for the transformer-encoder in the form of a programming language.
We map the basic components of a transformer-encoder---attention and feed-forward computation---into simple primitives, around which we form a programming language: the Restricted Access Sequence Processing Language (RASP). 
We show how RASP can be used to program solutions to tasks that could conceivably be learned by a Transformer, and how a Transformer can be trained to mimic a RASP solution. 
In particular, we provide RASP programs for histograms, sorting, and Dyck-languages. 
We further use our model to relate their difficulty in terms of the number of required layers and attention heads:
analyzing a RASP program implies a maximum number of heads and layers necessary to encode a task in a transformer.
Finally, we see how insights gained from our abstraction might be used to explain phenomena seen in recent works.
\end{abstract}

%% file: 01_introduction.tex
\section{Introduction}\label{Se:Intro}

We present a \emph{computational model for the transformer architecture} in the form of a simple language which we dub \rasp~(\emph{Restricted Access Sequence Processing Language}). Much as the token-by-token processing of RNNs can be conceptualized as finite state automata \cite{cleeremansRNNsDFAs}, our language captures the unique information-flow constraints under which a transformer operates as it processes input sequences. Our model helps reason about how a transformer operates at a higher-level of abstraction, reasoning in terms of a composition of \emph{sequence operations} rather than neural network primitives. 

\input{01a_acceptfig}

We are inspired by the use of automata as an abstract computational model for recurrent neural networks (RNNs). Using automata as an abstraction for RNNs has enabled a long line of work, including extraction of automata from RNNs \cite{NNExtraction,OurDFAExtraction,WFAfromRNNSpectral}, analysis of RNNs' practical expressive power in terms of automata \cite{LSTMsCanCount,RNNsasWFAs,MerrillAutomata,merrill-etal-2020-formal}, and even augmentations based on automata variants \cite{StackRNNsJoulin}. 
Previous work on transformers explores their computational power, but does not provide a computational model \cite{yun2019transformers,hahn2019limitations,perez2021transformersturing}.

Thinking in terms of the RASP model can help derive computational results. \pcite{bhattamishra2020ability} and \pcite{ebrahimi2020transformersdyck} explore the ability of transformers to recognize Dyck-k languages, with \citeauthor{bhattamishra2020ability} providing a construction by which Transformer-encoders can recognize a simplified variant of Dyck-$k$. Using \rasp, we succinctly express the construction of \cite{bhattamishra2020ability} as a short program, and further improve it to show, for the first time, that transformers can fully recognize Dyck-$k$ for all $k$.

Scaling up the complexity, \citet{clark2020transformers} showed empirically that transformer networks can learn to perform multi-step logical reasoning over first order logical formulas provided as input, resulting in ``soft theorem provers''. For this task, the mechanism of the computation remained elusive: how does a transformer perform even non-soft theorem proving?
As the famous saying by Richard Feynman goes, ``what I cannot create, I do not understand'': using {\rasp}, we were able to write a program that performs similar logical inferences over input expressions, and then ``compile'' it to the transformer hardware, defining a sequence of attention and multi-layer perceptron (MLP) operations. 

Considering computation problems and their implementations in {\rasp} allows us to ``think like a transformer'' while abstracting away the technical details of a neural network in favor of symbolic programs.
Recognizing that a task is representable in a transformer is as simple as finding a \rasp~ program for it, and
communicating this solution---previously done by presenting a hand-crafted transformer for the task---is now possible through a few lines of code. 
Thinking in terms of {\rasp} also allows us to shed light on a recent empirical observation of transformer variants \citep{sandwichtransformer}, and to find concrete limitations of ``efficient transformers'' with restricted attention \citep{tay2020efficient}.

In \Cref{Se:Experiments}, we show how a compiled {\rasp} program can indeed be realised in a neural transformer (as in \Cref{fig:accept}), and occasionally is even the solution found by a transformer trained on the task using gradient descent (Figs \ref{fig:histbos} and \ref{fig:reverse}).

{\bf Code} We provide a \rasp~ read-evaluate-print-loop (REPL) in §http://github.com/tech-srl/RASP§, along with a \rasp~ cheat sheet and link to replication code for our work.

%% file: 01a_acceptfig.tex
\begin{figure*}
    \begin{minipage}{.43\textwidth}
    \begin{lstlisting}[mathescape=true,escapechar=\%]
same_tok = select(tokens,tokens,==);
hist = selector_width(
            same_tok,
            assume_bos = True);
            % %
first = not has_prev(tokens);   
same_count = select(hist,hist,==);
same_count_reprs = same_count and
    select(first,True,==);
      % %
hist2 = selector_width(
            same_count_reprs,
            assume_bos = True);
\end{lstlisting}
    \end{minipage}
    \includegraphics[scale =0.25,trim = 270mm 140mm 000mm 200mm]{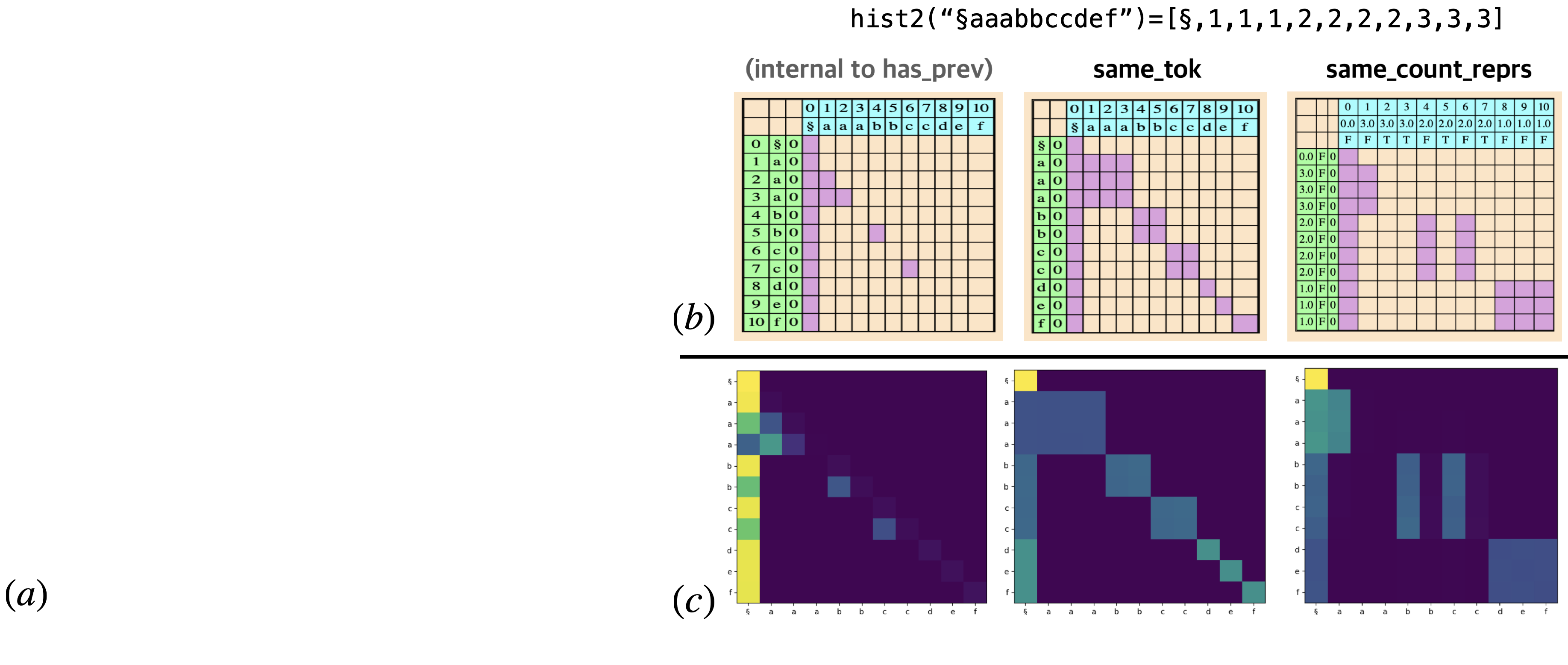}
    \caption{We consider \emph{double-histogram}, the task of counting for each input token how many unique input tokens have the same frequency as itself (e.g.: §hist2("\bos aaabbccdef")=[\bos,1,1,1,2,2,2,2,3,3,3]§). (a) shows a {\rasp} program for this task, 
    (b) shows the selection patterns of that same program, compiled to a transformer architecture and applied to the input sequence §\bos aaabbccdef§, 
    (c) shows the corresponding attention heatmaps, for the same input sequence, in a 2-layer 2-head transformer trained on double-histogram. This particular transformer was trained using both \emph{target} and \emph{attention} supervision, i.e.: in addition to the standard cross entropy loss on the target output, the model was given an MSE-loss on the difference between its attention heatmaps and those expected by the \rasp~ solution. The transformer reached test accuracy of $99.9\%$ on the task, and comparing the selection patterns in (b) with the heatmaps in (c) suggests that it has also successfully learned to replicate the solution described in (a).
    }
    \label{fig:accept}
\end{figure*}

%% file: 02_overview.tex
\section{Overview}\label{Se:Overview}

We begin with an informal overview of {\rasp}, with examples. The formal introduction is given in \Cref{Se:RASP}.

Intuitively, transformers' computations are applied to their entire input in parallel, using attention to draw on and combine tokens from several positions at a time as they make their calculations \cite{transformers,attentionbahdanau,attentionluong}. The iterative process of a transformer is then not along the length of the input sequence but rather the depth of the computation: the number of layers
it applies to its input as it works towards its final result. 

\noindent\textbf{The computational model.}\, Conceptually, a RASP computation over length-$n$ input involves manipulation of sequences of length $n$, and matrices of size $n\times n$. There are no sequences or matrices of different sizes in a RASP computation. The abstract computation model is as follows:

The input of a RASP computation is two sequences, \emph{tokens} and \emph{indices}. The first contains the user-provided input, and the second contains the range $0,1,...,n-1$.
The output of a RASP computation is a sequence, and the consumer of the output can choose to look only at specific output locations.

Sequences can be transformed into other sequences through element-wise operations. For example, for the sequences $s_1=[1,2,3]$ and $s_2=[4,5,6]$, we can derive $s_1+s_2=[5,7,9]$, $s_1+2=[3,4,5]$, $pow(s_1,2)=[1,4,9]$, $s_1>2=[F,F,T]$, $\text{pairwise\_mul}(s_1, s_2) = [4, 10, 18]$, and so on. 

\begin{figure}[t!]
\includegraphics[width=\linewidth, trim = 20mm 160mm 000mm -10mm ]{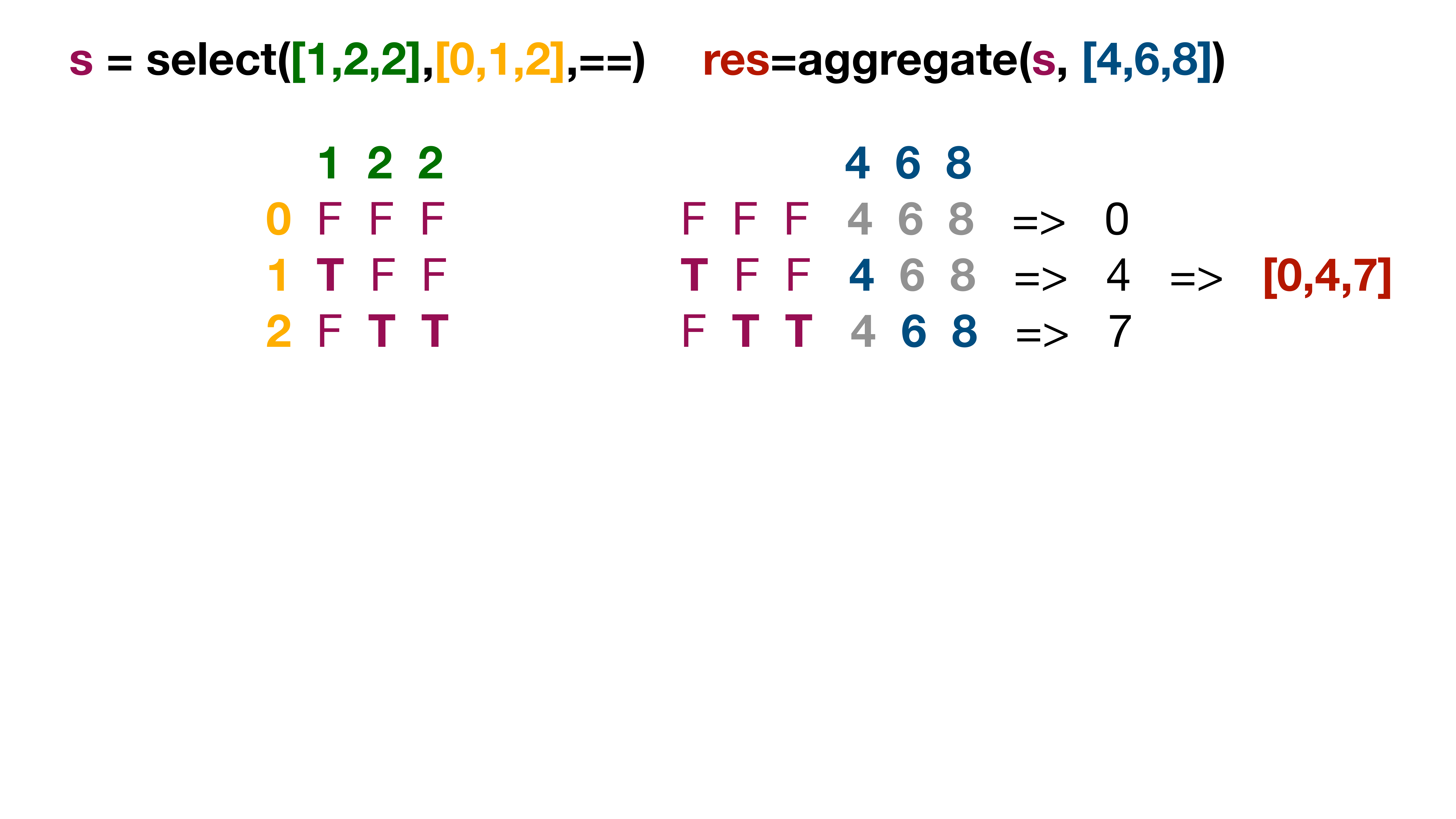}
\caption{Visualizing the select and aggregate operations. On the left, a selection matrix {\color{purple!60!black} \bf s} is computed by select, which marks for each {\color{orange} \bf query} position all of the {\color{green!40!black} \bf key} positions with matching values according to the given comparison operator §==§. On the right, aggregate uses {\color{purple!60!black} \bf s} as a filter over its input {\color{blue!60!black} \bf values}, averaging only the selected {\color{blue!60!black} \bf values} at each position in order to create its output, {\color{red!60!black} \bf res}. Where no {\color{blue!60!black} \bf values} have been selected, aggregate substitutes $0$ in its output.}
\vspace{-1em}
\label{fig:selagg}
\end{figure}

Sequences can also be transformed using a pair of \emph{select} and \emph{aggregate} operations (\Cref{fig:selagg}).
Select operations take two sequences $k,q$ and a boolean predicate $p$ over pairs of values, and return a \emph{selection matrix} $S$ such that for every $i,j\in[n]$, $S_{[i][j]}=p(k_{[i]}, q_{[j]})$. 
Aggregate operations take a matrix $S$ and a numeric sequence $v$, and return a sequence $s$ in which each position $s_{[i]}$ combines the values in $v$ according to row $i$ in $S$ (see full definition in \Cref{Se:RASP}).

Aggregate operations (over select matrices) are the only way to combine values from different sequence positions, or to move values from one position to another. For example, to perform the python computation:
§x = [a[0] for \_ in a]§,
we must first use $S$ = select$(indices,0,=$) to select the first position, and then $x$ = aggregate$(S,a)$ to broadcast it across a new sequence of the same length.

\smallparagraph{RASP programs are lazy functional,} and thus operate on functions rather than sequences. That is, instead of a sequence indices$=[0,1,2]$, we have a function §indices§ that returns $[0,1,2]$ on inputs of length $3$. Similarly, §s3=s1+s2§ is a function, that when applied to an input $x$ will produce the value §s3§$(x)$, which will be computed as §s1§$(x)+$§s2§$(x)$. We call these functions \emph{\encoders~ (\encoderfull)}. The same is true for the selection matrices, whose functions we refer to as \emph{selectors},
and the RASP language is defined in terms of \encoders~ and selectors, not sequences and matrices. However, the conceptual model to bear in mind is that of operations over sequences and selection matrices.

\smallparagraph{Example: Double Histograms} The \rasp~ program in \Cref{fig:accept} solves \emph{double-histogram}, the task of counting for each token how many unique input tokens in the sequence have the same frequency as its own: §hist2("\bos aabcd")=[\bos,1,1,3,3,3]§. 
The program begins by creating the 
the selector §same\_tok§, in which each input position focuses on all other positions containing the same token as its own,
and then applies the \rasp~ operation §selector\_width§ to it in order to obtain the \encoder~ §hist§, which computes the frequency of each token in the input: §hist("hello")=[1,1,2,2,1]§.
Next, the program uses the function §has\_prev§\footnote{Presented in \Cref{app:rasp:hist2} in \Cref{Se:app:task:sols}.} to create the \encoder~ §first§, which marks the first appearance of each token in a sequence: §first("hello")=[T,T,T,F,T]§. Finally, applying §selector\_width§ to the selector §same\_count\_reprs§, which focuses each position on all `first' tokens with the same frequency as its own, provides
§hist2§ as desired. 

\smallparagraph{Example: Shuffle-Dyck in {\rasp}} 
As an example of the kind of tasks that are natural to encode using {\rasp}, consider the Shuffle-Dyck language, in which multiple parentheses types must be balanced but do not have to satisfy any order with relation to each other. (For example, §"([)]"§ is considered balanced). In their work on transformer expressiveness, \pcite{bhattamishra2020ability} present a hand-crafted transformer for this language, including the details of which dimension represents which partial computation. 
{\rasp} can concisely describe the same solution, showing the high-level operations while abstracting away the details
of their arrangement into an actual transformer architecture.

\input{02a_shuffle_dyck}

We present this solution in \Cref{rasp:shuffledyck}:
the code 
compiles to a transformer architecture using 2 layers and a total of 3 heads,
exactly as in the construction of  \citeauthor{bhattamishra2020ability}. 
These numbers are inferred by the {\rasp} compiler: the programmer does not have to think about such details.

A pair of parentheses is balanced in a sequence if their running balance is never negative, and additionally is equal to exactly $0$ at the final input token. Lines 13--23 check this definition: lines 13 and 14 use §pair\_balance§ to compute the running balances of each parenthesis pair, and 17 checks whether these balances were negative anywhere in the sequence. The snippet in 21 (§bal1==0 and bal2==0§) creates an \encoder~ checking at each location whether both pairs are balanced, with the aggregation of line 20 loading the value of this \encoder~ from the last position. From there, a boolean composition of §end\_0§ and §had\_neg§ defines shuffle-dyck-$2$.

\smallparagraph{Compilation and Abstraction}
The high-level operations in {\rasp} can be compiled down to execute on a transformer: 
for example, the code presented in \Cref{fig:accept} compiles to a two-layer, $3$-head (total) architecture,
whose attention patterns when applied to the input sequence §"\bos aaabbccdef"§ are presented in \Cref{fig:accept}(b). 
(The full compiled computation flow for this program---showing how its component \encoders~ interact---is presented in  \Cref{Se:app:task:sols}).

{\rasp} abstracts away low-level operations into simple primitives, allowing a programmer to explore the full potential of a transformer
without getting bogged down in the details of how these are realized in practice. 
At the same time, 
{\rasp} enforces the information-flow constraints 
of
transformers,
preventing anyone from writing a program more powerful than they can express. 
One example of this is the lack of input-dependent loops in the \encoders, reflecting the fact that transformers cannot arbitrarily repeat operations\footnote{Though work exploring such transformer variants exists: \pcite{universalTransformers} devise a transformer architecture with a control unit, which can repeat its sublayers arbitrarily many times.}. 
Another is in the selectors: for each two positions, the decision whether one selects (`attends to') the other is pairwise.

We find {\rasp} a natural tool for conveying transformer solutions to given tasks.
It is modular and compositional, allowing us to focus on arbitrarily high-level computations when writing programs.
Of course, we are restricted to tasks for which a human \emph{can} encode a solution: we do not expect any researcher to implement, e.g., a strong language model or machine-translation system in {\rasp}---these are not realizable in any programming language. 
Rather, we focus on programs that convey concepts that people can encode in ``traditional'' programming languages, and the way they relate to the expressive power of the transformer.

In \Cref{Se:Experiments}, we will show empirically that {\rasp} solutions can indeed translate to real transformers. 
One example is given in \Cref{fig:accept}: having written a {\rasp} program (left) for the double-histograms task, we analyse it to obtain the number of layers and heads needed for a transformer to mimic our solution, and then train a transformer with supervision of both its outputs and its attention patterns to obtain a neural version of our solution (right). We find that the transformer can accurately learn the target attention patterns and use them to reach a high accuracy on the target task.

%% file: 02a_shuffle_dyck.tex
\begin{figure}
    \centering
    \begin{lstlisting}[mathescape=true,escapechar=\%]
def frac_prevs(sop,val){
  prevs = select(indices,indices,<=);
  return aggregate(prevs,
                indicator(sop==val));
}

def pair_balance(open,close) {
  opens = frac_prevs(tokens,open);
  closes = frac_prevs(tokens,close); 
  return opens - closes;
}
% %
bal1 = pair_balance("(",")"); 
bal2 = pair_balance("{","}");
% %
negative = bal1<0 or bal2<0;
had_neg = aggregate(select_all, 
                indicator(negative))>0;
select_last = select(indices,length-1,==);              
end_0 = aggregate(select_last, 
            bal1==0 and bal2==0);
    
shuffle_dyck2 = end_0 and not had_neg;
\end{lstlisting}
    \caption{{\rasp} program for the task shuffle-dyck-2 (balance 2 parenthesis pairs, independently of each other), capturing a higher level representation of the hand-crafted transformer presented by \pcite{bhattamishra2020ability}.
    \vspace{-1em}}
    \label{rasp:shuffledyck}
\end{figure}

%% file: 03_RASP.tex
\section{The {\rasp} language}\label{Se:RASP}
{\rasp} contains a small set of primitives and operations built around the core task of manipulating sequence processing functions referred to as \emph{\encoders~(\encoderfull)}, functions that take in an input sequence and return an output sequence of the same length. Excluding some atomic values, and the convenience of lists and dictionaries, \emph{everything} in {\rasp} is a function. Hence, to simplify presentation, we often demonstrate {\rasp} values with one or more input-output pairs: for example, §identity("hi")="hi"§\footnote{We use strings as shorthand for a sequence of characters.}. 

\rasp~ has a small set of built-in \encoders, and the goal of programming in \rasp~ is to compose these into a final \encoder~ computing the target task. For these compositions, the functions §select§ (creating selection matrices called \emph{selectors}), §aggregate§ (collapsing selectors and \encoders~ into a new \encoders), and §selector\_width§ (creating an \encoder~ from a selector) are provided, along with several elementwise operators reflecting the feed-forward sublayers of a transformer. As noted in \Cref{Se:Overview}, while all \encoders~ and selectors are in fact functions, we will prefer to talk in terms of the sequences and matrices that they create.
Constant values in \rasp~ (e.g., $2$, $T$, §h§) are treated as \encoders~ with a single value broadcast at all positions, and all symbolic values are assumed to have an underlying numerical representation which is the value being manipulated in practice.

\smallparagraph{The built-in \encoders} The simplest \encoder~ is the identity, given in {\rasp} under the name §tokens§: §tokens("hi")="hi"§. The other built-in \encoders~ are §indices§ and §length§, processing input sequences as their names suggest: §indices("hi")=[0,1]§, and §length("hi")=[2,2]§. 

\encoders~ can be combined with constants (numbers, booleans, or tokens) or each other to create new \encoders, in either an elementwise or more complicated fashion. 

\smallparagraph{Elementwise combination} 
of \encoders~ is done by the common operators for the values they contain, for example: §(indices+1)("hi")=[1,2]§, and §((indices+1)==length)("hi")=[F,T]§. This includes also a ternary operator:
§(tokens if (indices\%2==0) else "-")("hello")="h-l-o"§. 
When the condition of the operator is an \encoder~ itself, the result is an \encoder~ that is dependent on all 3 of the terms in the operator creating it.

\smallparagraph{Select and Aggregate} operations are used to combine information from different sequence positions. A selector takes two lists, representing \emph{keys} and \emph{queries} respectively, and a predicate $p$, and computes from these a selection matrix describing for each key, query pair $(k,q)$ whether the condition $p(k,q)$ holds. 

For example:
\begin{equation*}
sel([0,1,2],[1,2,3],<)=\begin{bmatrix}\mathbf{T} & F & F \\ \mathbf{T} & \mathbf{T} & F \\ \mathbf{T} & \mathbf{T} & \mathbf{T} \end{bmatrix}    
\end{equation*}
An aggregate operation takes one selection matrix and one list, and averages for each row of the matrix the values of the list in its selected columns. For example,
\vspace{-3pt}
\begin{equation*}
agg(\begin{bmatrix}\mathbf{T} & F & F \\ \mathbf{T} & \mathbf{T} & F \\ \mathbf{T} & \mathbf{T} & \mathbf{T} \end{bmatrix},[10,20,30])=[10,15,20]    
\end{equation*}
Intuitively, a select-aggregate pair can be thought of as a two-dimensional map-reduce operation. The selector can be viewed as performing filtering, and aggregate as performing a reduce operation over the filtered elements (see \Cref{fig:selagg}).

In \rasp, the selection operation is provided through the function §select§, which takes two \encoders~ §k§ and §q§ and a comparison operator $\circ$ and returns the composition of sel$(\cdot,\cdot,\circ)$ with §k§ and §q§, with this sequence-to-matrix function referred to as a \emph{selector}. For example: §a=select(indices,indices,<)§ is a selector, and §a("hey")=[[F,F,F],[T,F,F],[T,T,F]]§.
Similarly, the aggregation operation is provided through §aggregate§, which takes one selector and one \encoder~ and returns the composition of agg with these. For example: §aggregate(a,indices+1)("hey")=[0,1,1.5]§.\footnote{For convenience and efficiency, when averaging the filtered values in an aggregation, for every position where only a single value has been selected, {\rasp} passes that value directly to the output without attempting to `average' it. This saves the programmer from unnecessary conversion into and out of numerical representations when making simple transfers of tokens between locations: for example, using the selector §load1=select(indices,1,==)§, we may directly create the \encoder~ §aggregate(load1,tokens)("hey")="eee"§. Additionally, in positions when no values are selected, the aggregation simply returns a default value for the output (in \Cref{fig:selagg}, we see this with default value $0$), this value may be set as one of the inputs to the §aggregate§ function.}

\smallparagraph{Simple select-aggregate examples}
To create the \encoder~ that reverses any input sequence, we build a selector that requests for each query position the token at the opposite end of the sequence, and then aggregate that selector with the original input tokens: §flip=select(indices,length-indices-1,==)§, and §reverse=aggregate(flip,tokens)§. 
For example:
\vspace{-3pt}
$$\mathrm{§flip("hey")§}=\begin{bmatrix}F & F & \mathbf{T} \\ F & \mathbf{T} & F \\ \mathbf{T} & F & F \end{bmatrix} $$
$$\texttt{reverse("hey")}={\texttt{"yeh"}}$$
To compute the fraction of appearances of the token §"a"§ in our input, we build a selector that gathers information from \emph{all} input positions, and then aggregate it with a sequence broadcasting $1$ wherever the input token is §"a"§, and $0$ everywhere else. This is expressed as §select\_all=select(1,1,==)§,
and then §frac\_as = aggregate(select\_all,1 if tokens=="a" else 0)§.

\smallparagraph{Selector manipulations} Selectors can be combined elementwise using boolean logic. For example, for §load1=select(indices,1,==)§ and §flip§ from above:
\vspace{-5pt}
$$\mathrm{§(load1 or flip)("hey")§}=\begin{bmatrix}F & \mathbf{T} & \mathbf{T} \\ F & \mathbf{T} & F \\ \mathbf{T} & \mathbf{T} & F \end{bmatrix}$$
\smallparagraph{selector width} The final operation in \rasp~ is the powerful §selector\_width§, which takes as input a single selector and returns a new \encoder~ that computes, for each output position, the number of input values which that selector has chosen for it. This is best understood by example: using the selector §same\_token=select(tokens,tokens,==)§ that filters for each query position the keys with the same token as its own, we can compute its width to obtain a histogram of our input sequence: §selector\_width(same\_token)("hello")=[1,1,2,2,1]§.

\smallparagraph{Additional operations:} While the above operations are together sufficient to represent any {\rasp} program, {\rasp} further provides a library of primitives for common operations, such as §in§ -- either of a value within a sequence: §("i" in tokens)("hi")=[T,T]§, or of each value in a sequence within some static list: §tokens in ["a","b","c"])("hat")=[F,T,F]§.
\rasp~ also provides functions such as §count§, or §sort§.

\subsection{Relation to a Transformer}
We discuss how the {\rasp} operations compile to describe the information flow of a transformer architecture, suggesting how many heads and layers are needed to solve a task.

\smallparagraph{The built-in \encoders} §indices§ and §tokens§ reflect the initial input embeddings of a transformer, while §length§ is 
computed in \rasp:
§length= 1/aggregate(select\_all,indicator(indices==0))§, where §select\_all=select(1,1,==)§.

\smallparagraph{Elementwise Operations} reflect the feed-forward sub-layers of a transformer. 
These have overall not been restricted in any meaningful way: as famously shown by \pcite{horniketal1989}, MLPs such as those present in the feed-forward transformer sub-layers can approximate with arbitrary accuracy any borel-measurable function, provided sufficiently large input and hidden dimensions.

\smallparagraph{Selection and Aggregation}
Selectors translate to attention matrices, defining for each input the selection (attention) pattern used to mix the input values into a new output through weighted averages, and aggregation reflects this final averaging operation.
The uniform weights dictated by our selectors reflect an attention pattern in which `unselected' pairs are all given strongly negative scores, while the selected pairs all have higher, similar, scores. Such attention patterns are supported by the findings of \citep{merrill2020normgrowth}.

Decoupling selection and aggregation in {\rasp} allows §selectors§ to be reused in multiple aggregations, 
abstracting away the fact that these may actually require separate attention heads in the compiled architecture. Making selectors first class citizens also enables 
functions such as §selector\_width§, which take selectors as parameters. 

\smallparagraph{Additional abstractions}
All other operations, including the powerful §selector\_width§ operation, are implemented in terms of the above primitives. §selector\_width§ in particular can be implemented such that it compiles to either one or two selectors, depending on whether or not one can assume a beginning-of-sequence token is added to the input sequence. Its implementation is given in \Cref{Se:app:task:sols}.

\smallparagraph{Compilation} Converting an \encoder~ to a transformer architecture is as simple as tracing its computation flow out from the base \encoders. Each aggregation is an attention head, which must be placed at a layer later than all of its inputs. Elementwise operations are feedforward operations, and sit in the earliest layer containing all of their dependencies. Some optimisations are possible: for example, aggregations performed at the same layer with the same selector can be merged into the same attention head. A ``full" compilation---to concrete transformer weights---requires to e.g. derive MLP weights for the elementwise operations, and is beyond the scope of this work.
{\rasp} provides a method to visualize this compiled flow for any \encoder~ and input pair: Figures \ref{fig:reverse} and \ref{fig:histbos} were rendered using §draw(reverse,"abcde")§ and §draw(hist,"\bos aabbaabb")§.

%% file: 04_implications.tex
\vspace{-5pt}
\section{Implications and insights}\label{Se:Implications}
\smallparagraph{Restricted-Attention Transformers}
Multiple works propose restricting the attention mechanism to create more efficient transformers, reducing the time complexity of each layer from O$(n^2)$ to O$(nlog(n))$ or even O$(n)$ with respect to the input sequence length $n$ (see \citet{tay2020efficient} for a survey of such approaches). Several of these do so using \emph{sparse attention}, in which the attention is masked using different patterns to reduce the number of locations that can interact (\cite{child2019sparse,beltagy2020longformer,ainslie2020etc,zaheer2020bigbird,roy2020efficient}).

Considering such transformer variants in terms of {\rasp} allows us to reason about the computations they can and cannot perform. 
In particular, these variants of transformers all impose restrictions on the selectors, permanently forcing some of the $n^2$ index pairs in every selector to §False§. But does this necessarily weaken these transformers?

In \Cref{Se:app:task:sols} we present a sorting algorithm in {\rasp}, applicable to input sequences with arbitrary length and alphabet size\footnote{Of course, realizing this solution in real transformers requires sufficiently stable word and positional embeddings---a practical limitation that applies to all transformer variants.}. This problem is known to require at $\Omega(n\log(n))$ operations in the input length $n$---implying that a standard transformer can take full advantage of $\Omega(n\log(n))$ of the $n^2$ operations it performs in every attention head. It follows from this that all variants restricting their attention to $o(n\log(n))$ operations incur a real loss in expressive power.

\smallparagraph{Sandwich Transformers} 
Recently, \pcite{sandwichtransformer} showed that reordering the attention and feed-forward sublayers of a transformer affects its ability to learn language modeling tasks. 
In particular, they showed that: \begin{enumerate*}
\item pushing feed-forward sublayers towards the bottom of a transformer weakened it; and 
\item pushing attention sublayers to the bottom and feed-forward sublayers to the top strengthened it, provided there was still some interleaving in the middle. 
\end{enumerate*}

The base operations of {\rasp} help us understand the observations of \citeauthor{sandwichtransformer}. 
Any arrangement of a transformer's sublayers into a fixed architecture
imposes a restriction on the number and order of {\rasp} operations that can be chained in a program compilable to that architecture. 
For example, an architecture in which all feed-forward sublayers appear before the attention sublayers, imposes that no elementwise operations may be applied to the result of any aggregation.

In {\rasp}, there is little value to repeated elementwise operations before the first §aggregate§: each position has only its initial input, and cannot generate new information. This explains the first observation of \citeauthor{sandwichtransformer}.
In contrast, an architecture beginning with several attention sublayers---i.e., multiple §select-aggregate§ pairs---will be able to gather a large amount of information into each position early in the computation,
even if only by simple rules\footnote{While the attention sublayer of a transformer does do some local manipulations on its input to create the candidate output vectors, it does not contain the powerful MLP with hidden layer as is present in the feed-forward sublayer.}.
More complicated gathering rules can later be realised by applying elementwise operations to aggregated information before generating new selectors,
explaining the second observation.

\smallparagraph{Recognising Dyck-$k$ Languages}
The Dyck-$k$ languages---the languages of sequences of correctly balanced parentheses, with $k$ parenthesis types---have been heavily used in considering the expressive power of RNNs \cite{sennhauser2018lstmsCFGs,skachkova-etal-2018-closing,bernardy-2018-recurrent,MerrillAutomata,hewitt2020rnnsdyck}.

Such investigations motivate similar questions for transformers, and several works approach the task.
\pcite{hahn2019limitations} proves that transformer-encoders with hard attention cannot recognise Dyck-$2$. 
\pcite{bhattamishra2020ability} and \pcite{yao2021transformersdyckkm} provide transformer-encoder constructions for recognizing simplified variants of Dyck-$k$,
though the simplifications are such that no conclusion can be drawn for unbounded depth Dyck-$k$ with $k>1$.
Optimistically, \pcite{ebrahimi2020transformersdyck} train a transformer-encoder with causal attention masking 
to process Dyck-$k$ languages with reasonable accuracy for several $k>1$,
finding that it learns a stack-like behaviour to complete the task.

We consider Dyck-$k$ using \rasp, specifically defining Dyck-$k$-PTF as the task of classifying for every prefix of a sequence whether it is legal, but not yet balanced ({\bf P}), balanced ({\bf T}), or illegal ({\bf F}). We show that \rasp~ can solve this task in a fixed number of heads and layers for \emph{any} $k$,  presenting our solution in \Cref{Se:app:task:sols}\footnote{We note that \rasp~ does not suggest the embedding width needed to encode this solution in an actual transformer.}.

\smallparagraph{Symbolic Reasoning in Transformers}
\pcite{clark2020transformers} show that transformers are able to emulate symbolic reasoning: they train a transformer which, given the facts ``Ben is a bird" and ``birds can fly", correctly validates that ``Ben can fly". Moreover, they show that transformers are able to perform several logical `steps': given also the fact that only winged animals can fly, their transformer confirms that Ben has wings.
This finding however does not shed any light on \emph{how} the transformer is achieving such a feat.

{\rasp} empowers us to approach the problem on a high level. We write a {\rasp} program for the related but simplified problem of containment and inference over sets of elements, sets, and logical symbols, in which the example is written as §b§$\in$§B, x$\in$B$\rightarrow$x$\in$F, b§$\in$§F?§ (implementation available in our repository). 
The main idea is to store at the position of each set symbol the elements contained and not contained in that set, and at each element symbol the sets it is and is not contained in. Logical inferences are computed by passing information between symbols in the same `fact', and propagated through pairs of identical set or element symbols, which share their stored information.

\smallparagraph{Use of Separator Tokens}
\pcite{clark2019bertattention} observe that many attention heads in BERT \cite{BERT2018orig} (sometimes) focus on separator tokens, speculating that these are used for ``no-ops" in the computation. \cite{ebrahimi2020transformersdyck} find that transformers more successfully learn Dyck-$k$ languages when the input is additionally provided with a beginning-of-sequence (BOS) token, with the trained models treating it as a base in their stack when there are no open parentheses.
Our {\rasp} programs suggest an additional role that such separators may be playing: by providing a fixed signal from a `neutral' position, separators facilitate conditioned counting in transformers, that use the diffusion of the signal to compute how many positions a head was attending to. Without such neutral positions, counting requires an additional head, such that an agreed-upon position may artificially be treated as neutral in one head and then independently accounted for in the other.

A simple example of this is seen in \Cref{fig:histbos}. There, §selector\_width§ is applied with a BOS token, creating in the process an attention pattern that focuses on the first input position (the BOS location) from all query positions, in addition to the actual positions selected by §select(tokens,tokens,==)§. A full description of §selector\_width§ is given in \Cref{Se:app:task:sols}.

%% file: 05_experiments.tex
\vspace{-5pt}
\section{Experiments}\label{Se:Experiments}
We evaluate the relation of {\rasp} to transformers on three fronts: \begin{enumerate*}
    \item its ability to upper bound the number of heads and layers required to solve a task,
    \item the tightness of that bound,
    \item its feasibility in a transformer, i.e., whether a sufficiently large transformer can encode a given {\rasp} solution.
\end{enumerate*}, training several transformers.  We relegate the exact details of the transformers and their training to \Cref{Se:app:experiments}.

For this section, we consider the following tasks:
\begin{compactenum}
    \item Reverse, e.g.: §reverse("abc")="cba"§.
    \item Histograms, with a unique beginning-of-sequence (BOS) token §\bos§ (e.g., §hist\_bos("\bos aba")=[\bos,2,1,2]§) and without it (e.g.,  §hist\_nobos("aba")=[2,1,2]§).
    \item Double-Histograms, with BOS: for each token, the number of unique tokens with same histogram value as itself. E.g.: §hist2("\bos abbc")=[\bos,2,1,1,2]§. 
    \item Sort, with BOS: ordering the input tokens lexicographically. e.g.: §sort("\bos cba")="\bos abc"§.
    \item Most-Freq, with BOS: returning the unique input tokens in order of decreasing frequency, with original position as a tie-breaker and the BOS token for padding. E.g.: §most\_freq("\bos abbccddd")="\bos dbca\bos\bos\bos\bos "§.
    \item Dyck-$i$ PTF, for $i=1,2$: the task of returning, at each output position, whether the input prefix up to and including that position is a legal Dyck-$i$ sequence (§T§), and if not, whether it can (§P§) or cannot (§F§) be continued into a legal Dyck-$i$ sequence. E.g: §Dyck1\_ptf("()())")="PTPTF"§.
\end{compactenum}
We refer to double-histogram as 2-hist, and to each Dyck-$i$ PTF problem simply as Dyck-$i$.
The full {\rasp} programs for these tasks, and the computation flows they compile down to, are presented in \Cref{Se:app:task:sols}. The size of the transformer architecture each task compiles to is presented in \Cref{tab:sizes}.

\input{05a_sizes}

\smallparagraph{Upper bounding the difficulty of a task}\label{Se:exp:upper}
Given a {\rasp} program for a task, e.g. double-histogram as described in \Cref{fig:accept}, we can compile it down to a transformer architecture, effectively predicting the maximum number of layers and layer width (number of heads in a layer) needed to solve that task in a transformer. To evaluate whether this bound is truly sufficient for the transformer, we train $4$ transformers of the prescribed sizes on each of the tasks.

\input{05bii_reverse}

We report the accuracy of the best trained transformer for each task in \Cref{tab:sizes}. Most of these transformers reached accuracies of $99.5\%$ and over, suggesting that the upper bounds obtained by our programs are indeed sufficient for solving these tasks in transformers. For some of the tasks, we even find that the {\rasp} program is the same as or very similar to the `natural' solution found by the trained transformer. In particular, Figures \ref{fig:reverse} and \ref{fig:histbos} show a strong similarity between the compiled and learned attention patterns for the tasks Reverse and Histogram-BOS, 
though the transformer trained on Reverse appears to have learned a different mechanism for computing §length§ than that given in {\rasp}.

\input{05bi_hist}

\smallparagraph{Tightness of the bound}\label{Se:exp:tight}
We evaluate the tightness of our {\rasp} programs by training smaller transformers than those predicted by our compilation, and observing the drop-off in test accuracy. Specifically, we repeat our above experiments, but this time we also train each task on up to 4 different sizes. In particular, denoting $L,H$ the number of layers and heads predicted by our compiled {\rasp} programs, we train for each task transformers with sizes $(L,H)$, $(L-1,H)$, $(L,H-1)$, and $(L-1,2H)$ (where possible) \footnote{The transformers of size $(L-1,2H)$ are used to validate that any drop in accuracy is indeed due to the reduction in number of layers, as opposed to the reduction in total heads that this entails. 
However, doubling $H$ means the embedding dimension will be divided over twice as many heads. To counteract any negative effect this may have, we also double the embedding dimension for these transformers.}.

\input{05c_dropoff}

We report the average test accuracy reached by each of these architectures in \Cref{tab:dropoff}.
For most of the tasks, the results show a clear drop in accuracy as the number of heads or layers is reduced below that obtained by our compiled {\rasp} solutions for the same tasks---several of these reduced transformers fail completely to learn their target languages. 

The main exception to this is sort, which appears unaffected by the removal of one layer, and even achieves its best results in this case. 
Drawing the attention pattern for the single-layer sort transformers reveals relatively uniform attention patterns.
It appears that the transformer has learned to take advantage of the bounded input alphabet size, effectively implementing bucket sort for its task.
This is because a single full-attention head is sufficient to compute for every token its total appearances in the input, from which the correct output can be computed locally at every position.

\smallparagraph{Feasibility of a {\rasp} program}\label{Se:exp:force}
We verify that a given {\rasp} program can indeed be represented in a transformer. For this, we return to the tougher tasks above, and this time train the transformer with an additional loss component encouraging it to learn the attention patterns created in our compiled solution (i.e., we supervise the attention patterns in addition to the target output). In particular, we consider the tasks double-histogram, sort, and most-freq, all with the assumption of a BOS token in the input. After training each transformer for $250$ epochs with both target and attention supervision, they all obtain high test accuracies on the task ($99{+}\%$), and appear to encode attention patterns similar to those compiled from our solutions. We present the obtained patterns for double-histogram, alongside the compiled {\rasp} solution, in \Cref{fig:accept}. We present its full computation flow, as well as the learned attention patterns and full flow of sort and most-freq, in \Cref{Se:app:experiments}.

%% file: 05a_sizes.tex
\begin{table}[t]
\small 
    \centering
    \begin{tabular}{c||c|c|c|c}
    \toprule 
        & & & & Attn.  \\
        Language & Layers & Heads & Test Acc. & Matches? \\
        
    \midrule
        Reverse & $2$ & $1$ & $99.99\%$ & \halfcheck \\
        Hist BOS & $1$ & $1$ & $100\%$ & \checkmark \\
        Hist no BOS & $1$ & $2$ & $99.97\%$ & \halfcheck \\
        Double Hist & $2$ & $2$ & $99.58\%$ & \halfcheck \\
        Sort & $2$ & $1$ & $99.96\%$ & \xmark \\
        Most Freq & $3$ & $2$ & $95.99\%$ & \xmark \\
        Dyck-$1$ PTF & $2$ & $1$ & $99.67\%$ & \halfcheck \\
        Dyck-$2$ PTF \footnotemark & $3$ & $1$ & $99.85\%$ & \xmark \\
      \bottomrule
    \end{tabular}
    \caption{Does a {\rasp} program correctly upper bound the number of heads and layers needed for a transformer to solve a task? In the left columns, we show the compilation size of our {\rasp} programs for each considered task, and in the right columns we show the best (of 4) accuracies of transformers trained on these same tasks, and evaluate whether their attention mechanisms appear to match (using a \halfcheck for partially similar patterns: see \Cref{fig:reverse} for an example). For {\rasp} programs compiling to varying number of heads per layer, we report the maximum of these.
    \vspace{-1em}
    }
    \label{tab:sizes}
\end{table}

\footnotetext{The actual optimal solution for Dyck-$2$ PTF cannot be realised in {\rasp} as is, as it requires the addition of a select-\emph{best} operator to the language---reflecting the power afforded by softmax in the transformer's self-attention. In this paper, we always refer to our analysis of Dyck-$2$ with respect to this additional operation.}

%% file: 05bii_reverse.tex
\begin{figure}[t!]
\centering
    \begin{minipage}{1\textwidth}
    \begin{lstlisting}[mathescape=true,escapechar=\%]
opp_index = length - indices - 1;
flip = select(indices, opp_index,==);
reverse = aggregate(flip, tokens);
\end{lstlisting}
    \end{minipage}
    \includegraphics[scale =0.25,trim = 0 80 50 0] {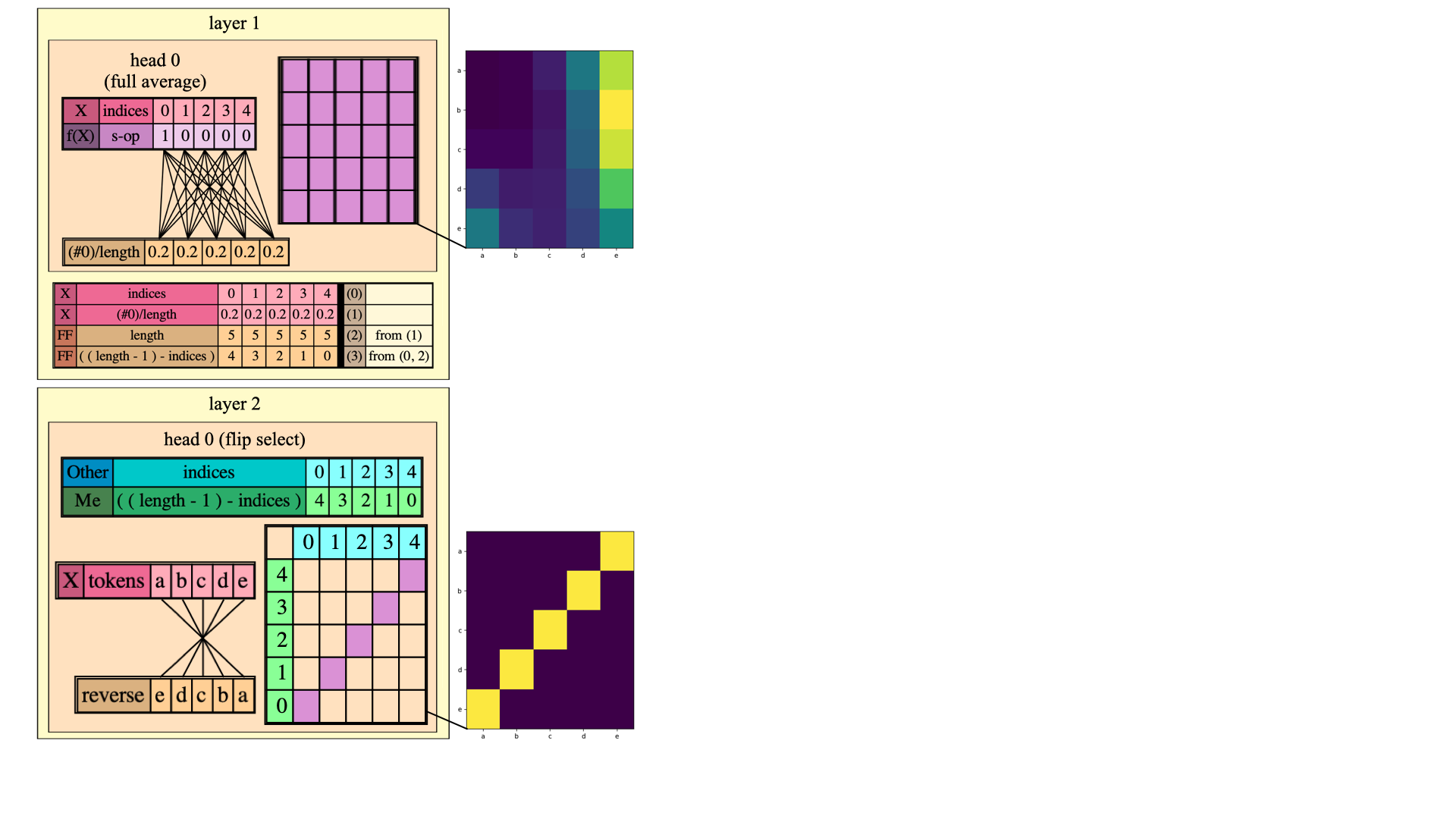}
\caption{Top: {\rasp} code for computing reverse (e.g., §reverse("abc")="cba"§). Below, its compilation to a transformer architecture (left, obtained through §draw(reverse,"abcde")§ in the \rasp~ REPL), and the attention heatmaps of a transformer trained on the same task (right), both visualised on the same input. 
Visually, the attention head in the second layer of this transformer corresponds perfectly to the behavior of the §flip§ selector described in the program. The head in the first layer, however, appears to have learned a different solution from our own: instead of focusing uniformly on the entire sequence (as is done in the computation of §length§ in \rasp), this head shows a preference for the last position in the sequence.
\vspace{-1em}
}
    \label{fig:reverse}
\end{figure}

%% file: 05bi_hist.tex
\begin{figure*}[t]
\centering
\hspace{-80pt}
    \begin{minipage}{0.45\textwidth}
    \begin{lstlisting}[mathescape=true,escapechar=\%,numbers=left] 
same_tok = select(tokens, tokens, ==);
hist = selector_width(same_tok,
            assume_bos = True);
\end{lstlisting}
    \end{minipage}  
\hspace{-50pt}
\begin{minipage}{0.5\textwidth}
    \includegraphics[scale=0.17,trim=0 0 0 0 ]{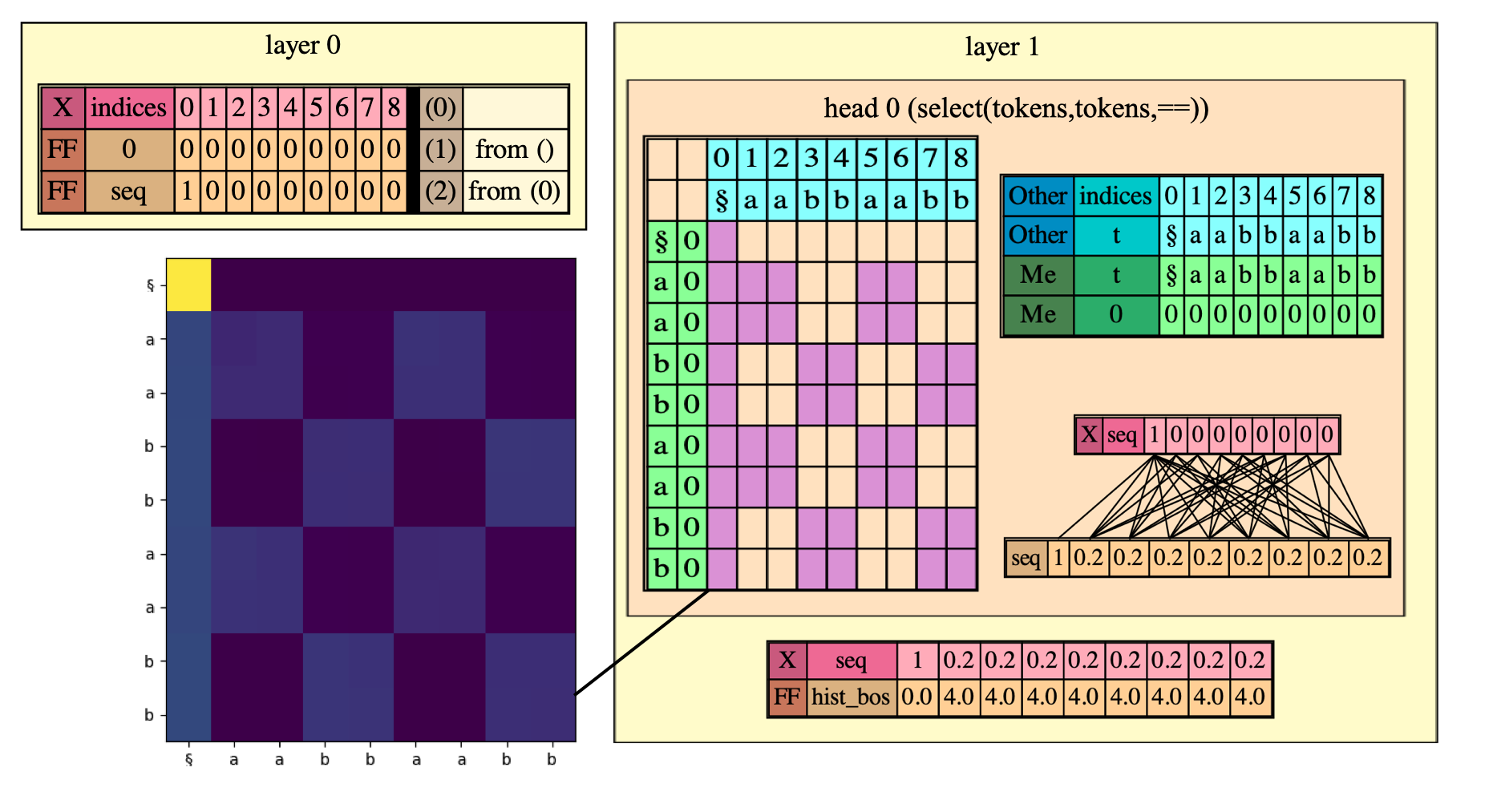}
\end{minipage}  
    \caption{The {\rasp} program for computing with-BOS histograms (left),
    alongside its compilation to a transformer architecture (cream boxes) and the attention head (center bottom) of a transformer trained on the same task, without attention supervision.
    The compiled architecture and the trained head are both presented on the same input sequence, §"\bos aabbaabb"§. The transformer architecture was generated in the RASP REPL using §draw(hist,"\bos aabbaabb")§. 
    }

    \label{fig:histbos}
\end{figure*}

%% file: 05c_dropoff.tex
\begin{table}[t]
\small 
    \centering
    \begin{tabular}{c||c|c|c|c|c}
    \toprule 
        Language & {\rasp} & \multicolumn{4}{c}{Average test accuracy ($\%$) with...} \\
        \cline{3-6}
        &  $L,H$ & $L,H$ & $H{-}1$ & $L{-}1$ & $L{-}1,2H$ \tstrut \\
        
    \midrule
        Reverse & $2,1$ & $\mathbf{99.9}$ & - & $23.1$ & $41.2$ \\
        Hist & $1,2$ & $\mathbf{99.9}$ & $91.9$ & - & - \\
        2-Hist & $2,2$ & $\mathbf{99.0}$ & $73.5$ & $40.5$ & $83.5$ \\
        Sort & $2,1$ & $99.8$ & - & $99.0$ & $\mathbf{99.9}$  \\
        Most Freq & $3,2$ & $\mathbf{93.9}$ & $92.1$ & $84.0$ & $90.2$ \\
        Dyck-$1$ & $2,1$ & $\mathbf{99.3}$ & - &$96.9$ & $96.4$ \\
        Dyck-$2$ & $3,1$ & $\mathbf{99.7}$ & - & $98.8$ & $94.1$ \\
      \bottomrule
    \end{tabular}
    \caption{Accuracy dropoff in transformers
    when reducing their number of heads and layers relative to the compiled {\rasp} solutions for the same tasks. 
    The transformers trained on the size predicted by {\rasp} have very high accuracy, and in most cases there is a clear drop as that size is reduced.
    Cases creating an impossible architecture ($H$ or $L$ zero) are marked with -. Histogram with BOS uses only 1 layer and 1 head, and so is not included.
    As in \Cref{tab:sizes}, Dyck-$2$ is considered with the addition of §select\_best§ to {\rasp}. 
    \vspace{-1em}
     }
    \label{tab:dropoff}
\end{table}

%% file: 06_conclusions.tex
\vspace{-5pt}
\section{Conclusions}\label{Se:Conclusions}
We abstract the computation model of the transformer-encoder as a simple sequence processing language, {\rasp}, that captures the unique constraints on information flow present in a transformer. Considering computation problems and their implementation in {\rasp} allows us to ``think like a transformer'' while abstracting away the technical details of a neural network in favor of symbolic programs. We can analyze any {\rasp} program to infer the minimum number of layers and maximum number of heads required to realise it in a transformer.
We show several examples of programs written in the {\rasp} language, showing how operations can be implemented by a transformer,
and train several transformers on these tasks, finding that RASP helps predict the number of transformer heads and layers needed to solve them.
Additionally, we use {\rasp} to shed light on an empirical observation over transformer variants,
and find concrete limitations for some ``efficient transformers''.

%% file: App00_all.tex
\clearpage

\begin{figure*}[!t]
\centering
\includegraphics[scale =0.36,trim = 0 0 800 0] {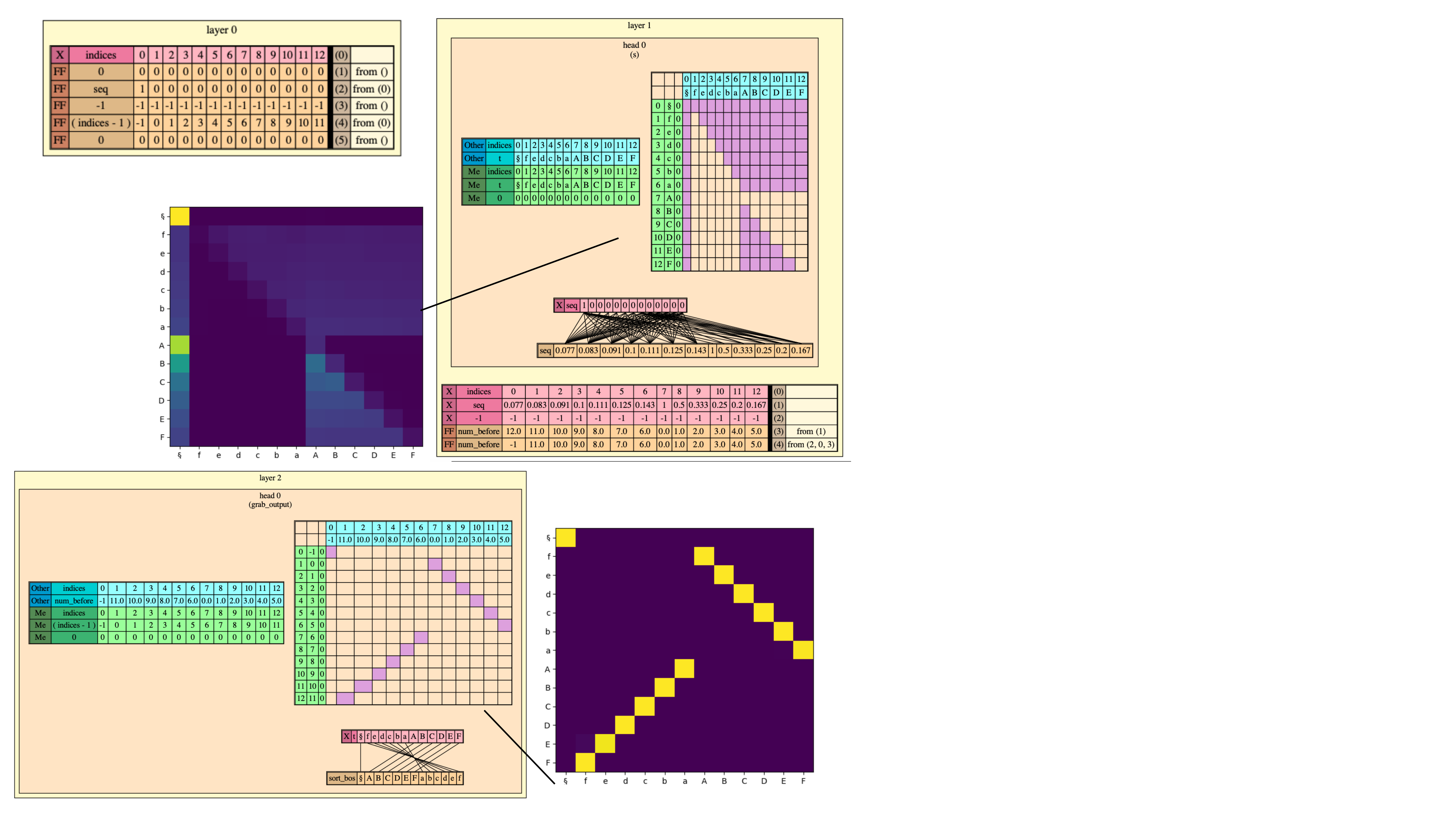}
\caption{Computation flow in compiled architecture from {\rasp} solution for sort (with BOS token), alongside heatmaps from the corresponding heads in a transformer trained with both target and attention supervision on the same task and \rasp~ solution. The \rasp~ solution is simply written §sort(tokens,tokens,assume\_bos=True)§, using the function §sort§ shown in \Cref{app:rasp:dyck1}. Both the {\rasp} architecture and the transformer are applied to the input sequence §"\bos fedcbaABCDEF"§.}\label{app:fig:sort}
\end{figure*}

\appendix

\section*{Appendices}

In \Cref{Se:app:experiments} we give training details from the experiments in this paper, as well as additional results from the transformers trained to mimic {\rasp}-predicted attention patterns. 
The exact {\rasp} solutions for all tasks considered in the paper, as well as an implementation of the operation §selector\_width§ in terms of other operations (which have direct translation to a transformer), are presented in \Cref{Se:app:task:sols}. This section also presents the computation flows in compiled architectures for several of these solutions.

\input{App3_experiments}

\input{App2_task_solutions}

%% file: App3_experiments.tex
\section{Experiments}\label{Se:app:experiments}

\subsection{Results: Attention-regularised transformers}
We trained $3$ transformers with a target attention pattern according to our {\rasp} solutions, these $3$ being for the tasks double-histogram, sort, and most-freq as described in the paper.
All of these reached high ($99{+}\%$) accuracy on their sequence-to-sequence task, computed as fraction of output tokens predicted correctly.
Plotting their attention patterns also shows clear similarity to those of the compiled {\rasp} programs:

For the \emph{double-histogram} task, a full compiled architecture is presented on the sequence §\bos aabbaa§ in \Cref{fig:hist2_full}. Additionally, in \Cref{fig:accept}, just its attention patterns are presented alongside the corresponding attention heads from its attention-regularised transformer, this time both on the sequence §\bos aabbaabb§. 

For the \emph{sorting} task, we present a full computation flow on the input sequence §\bos fedcbaABCDEF§, alongside the corresponding attention heads of the regularised transformer on the same sequence, in \Cref{app:fig:sort}. The regularised transformer had input alphabet of size $52$ and reached test accuracy $99.0\%$ on the task (measured as percentage of output positions where the correct output token had the maximum probability).

\begin{figure*}
\centering
    \includegraphics[scale =0.5,trim = 0 0 1200 0 ] {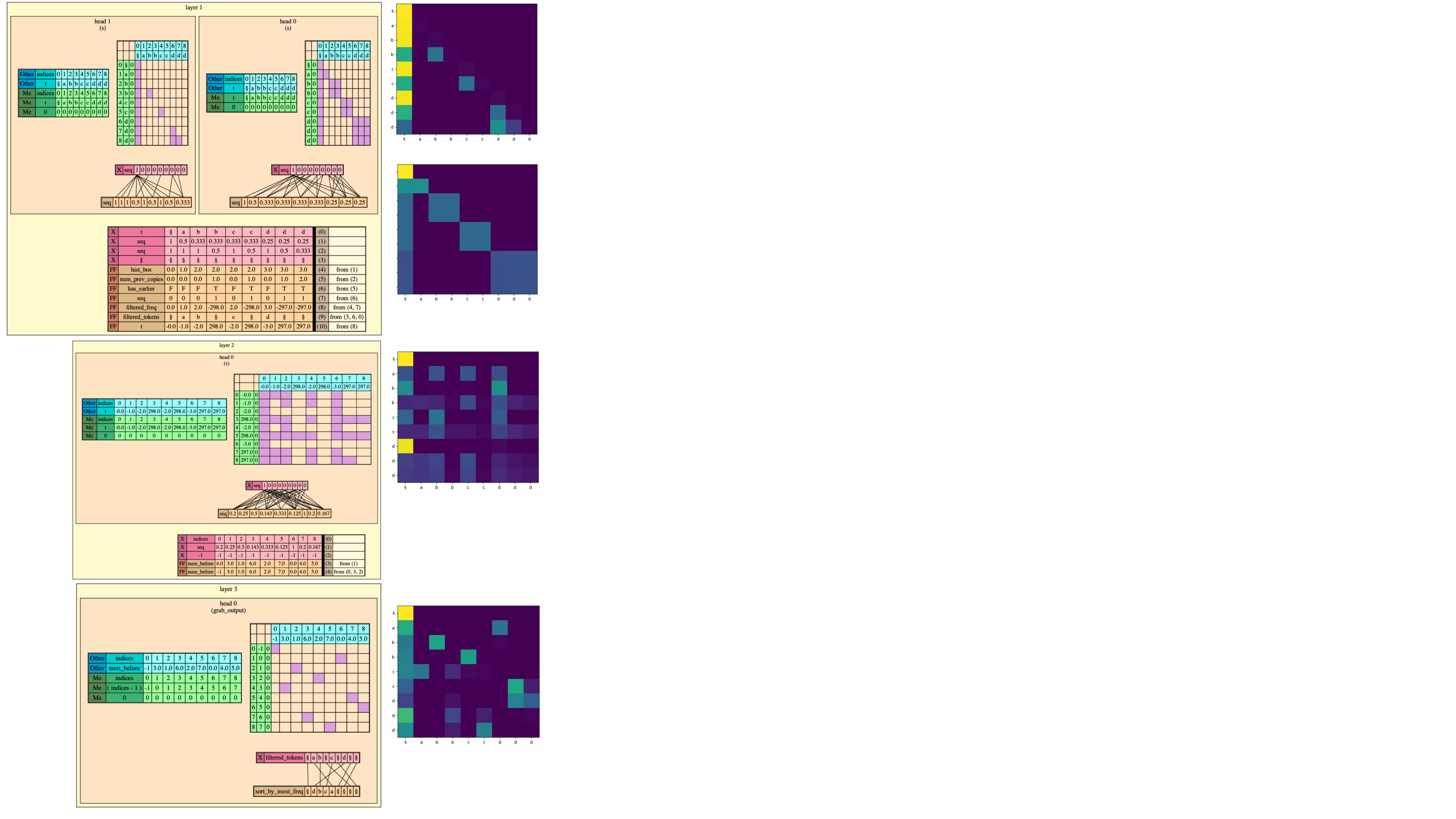}
\caption{Computation flow in compiled architecture from {\rasp} solution for sorting by frequency (returning all unique tokens in an input sequence, sorted by decreasing frequency), alongside heatmaps from attention heads in transformer trained on same task and regularised to create same attention patterns. Both are presented on the input sequence §\bos abbccddd§, for which the correct output is §\bos dbca§. The transformer architecture has 3 layers with 2 heads apiece, but the {\rasp} architecture requires only 1 head for each of the second and third layers. We regularised only one for each of these and present just that head.
    }
    \label{app:fig:by_freq}
\end{figure*}

For the \emph{most-freq} task (returning each unique token in the input, by descending order of frequency, and padding the rest with the BOS token) we do the again show a computation flow alongside the regularised transformer, this time in \Cref{app:fig:by_freq} and with the sequence §\bos aabbcddd§. On this task the regularised transformer had input alphabet of size $26$ and reached test accuracy $99.9\%$.

\subsection{Training Details}

In the upper bound and tightness experiments (\Cref{Se:Experiments}),
for each task and layer/head specification, we train transformers with embedding dimension $256$ and feed-forward dimension $512$ on the task for $100$ epochs. We use learning rates $0.0003$ and $0.0001$, and learning rate decay $\gamma=0.98$ and $0.99$, training $4$ transformers overall for each task. We use the ADAM optimiser and no dropout. Each transformer is trained on sequences of length $0--100$, with train/validation/test set sizes of $50,000$, $1,000$, and $1,000$ respectively. Excluding the BOS token, the alphabet sizes are: $3$ and $5$ and for Dyck-$1$ and Dyck-$2$ (the parentheses, plus one neutral token), $100$ for reverse and sort, and $26$ for the rest (to allow for sufficient repetition of tokens in the input sequences). All input sequences are sampled uniformly from the input alphabet and length, with exception of the Dyck languages, for which they are generated with a bias towards legal prefixes to avoid most outputs being §F§.

For the attention regularised transformers, we make the following changes: first, we only train one transformer per language, with learning rate $0.0003$ and decay $0.98$. We train each transformer for $250$ epochs (though they reach high validation accuracy much earlier than that). The loss this time is added to an MSE-loss component, computed from the differences between each attention distribution and its expected pattern. As this loss is quite small, we scale it by a factor of $100$ before adding it to the standard output loss.

%% file: App2_task_solutions.tex
\section{{\rasp} programs and computation flows for the tasks considered}\label{Se:app:task:sols}

\subsection{§selector\_width§} The {\rasp} implementation of §selector\_width§ is presented in \Cref{fig:selector_width}.
 The core observation is that, by using a selector that always focuses on zero (§or0§ in the presented code), we can compute the inverse of that selector's width by aggregating a $1$ from position $0$ and $0$ from everywhere else. It then remains only to make a correction according to whether or not the selector was actually focused on $0$, using the second selector §and0§ (if there isn't a beginning-of-sequence token) or our prior knowledge about the input (if there is).

\subsection{{\rasp} solutions for the paper tasks}
We now present the {\rasp} solutions for each of the tasks considered in the paper, as well as an implementation of the {\rasp} primitive §selector\_width§ in terms of only the primitives §select§ and §aggregate§. 

The solution for histograms, with or without a BOS token, is given in \Cref{app:rasp:hist}. The code for double-histograms (e.g., §hist2("§aaabbccdef")=[§1,1,1,2,2,2,2,3,3,3]§) is given in \Cref{app:rasp:hist2}. The general sorting algorithm (sorting any one sequence by the values (`keys') of any other sequence) is given in \Cref{app:rasp:sort}, and sorting the tokens by their frequency ("Most freq") is given in \Cref{app:rasp:by_freq}. Descriptions of these solutions are in their captions. 

\paragraph{The Dyck-PTF Languages}

\emph{Dyck-$1$-PTF} First each position attends to all previous positions up to and including itself in order to compute the balance between opening and closing braces up to itself, not yet considering the internal ordering of these. Next, each position again attends to all previous positions, this time to see if the ordering was problematic at some point (i.e., there was a negative balance). From there it is possible to infer for each prefix whether it is balanced (T), could be balanced with some more closing parentheses (P), or can no longer be balanced (F). We present the code in \Cref{app:rasp:dyck1}.

\input{App2b_dyck2}

\emph{Dyck-$2$-PTF}
For this descripition we differentiate between instances of an opening and closing parenthesis (\emph{opener} and \emph{closer}) matching each other with respect to their position within a given sequence, e.g. as §(,>§ and §\{,]§ do in the sequence §(\{]>§, and of the actual tokens matching with respect to the pair definitions, e.g. as the token pairs §\{,\}§ and §(,)§ are defined. For clarity, we refer to these as structure-match and pair-match, respectively. 

For a Dyck-$n$ sequence to be balanced, it must satisfy the balance checks as described in Dyck-$1$ (when treating all openers and all closers as the same), and additionally, it must satisfy that every structure-matched pair is also a pair-match. 

We begin by using the function §num\_prevs§ from \Cref{app:rasp:dyck1} to compute balances as for Dyck-$1$, ignoring which token pair each opener or closer belongs to. Next, we create an attention pattern §open\_for\_close§ that focuses each closer on its structure-matched opener, and use that pattern to pull up the structure-matched opener for each closer (the behaviour of that pattern on closers that do not have structure-matched openers is not important: in this case there will anyway be a negative balance at that closer). For each location, we then check that it does not have an earlier negative balance, and it does not have an earlier closer whose structure-matched opener is not a pair-match. If it fails these conditions the output is F, otherwise it is T if the current balance is 0 and P otherwise. The remaining challenge is in computing §open\_for\_close§.

In pure {\rasp}---i.e., within the language as presented in this work---this is realisable in two steps. First, we number each parenthesis according to how many previous parentheses have the same depth as itself, taking for openers the depth after their appearance and for closers the depth before. For example, for §(())()§, the depths are §[1,2,2,1,1,1]§, and the depth-index is §[1,1,2,2,3,3]§. Then, each closer's structure-matched opener is the opener with the same depth as itself, and depth-number immediately preceding its own. This solution is given in \Cref{app:rasp:dyck2}, and compiles to 4 layers with maximum width 2.

However, by adding the theoretical operation §select\_best§, and a scorer object similar to selectors (with number values as opposed to booleans), we can reduce the computation of §open\_for\_close§ to simply: ``find the last opener with the same depth as the closer, that is still before the closer''. In this case, the depth-index of each position does not need to be computed in order to obtain §open\_for\_close§, saving the layer and 2 heads that its compilation creates. 
We now elaborate on §select\_best§ and this alternative computation of §open\_for\_close§.

\smallparagraph{§select\_best§ (Theoretical)}
First, we imagine a new operator §score§ that behaves similarly to §select§, except that: instead of using predicates such as §==, <§, or §>§ to compare values and create \emph{selectors}, it expects the values to be numbers and simply multiplies them to create \emph{scorers}. For example (presented on concrete input for clarity, similar to the presentations in \Cref{Se:RASP}):
\begin{equation*}
\mathrm{score}([0,1,2],[1,-1,1])=\begin{bmatrix} 0 & 1 & 2 \\ 0 & -1 & -2 \\ 0 & 1 & 2  \end{bmatrix}    
\end{equation*}

Next we define §select\_best§ as a function taking one selector $sel$ and one scorer $sc$, and returning a new selector in which, for each position, only the selected value of $sel$ with the highest score in $sc$ remains. For example (presented again on concrete input for clarity, as opposed to function-building syntax of RASP):
\begin{equation*}
\mathrm{sel\textunderscore best}
(\begin{bmatrix} \mathbf{T} & \mathbf{T} & \mathbf{T} \\ \mathbf{T} & \mathbf{T} & F \\ F & F & F  \end{bmatrix}, 
\begin{bmatrix} 0 & 1 & 2 \\ 0 & 1 & 2 \\ 0 & 1 & 2  \end{bmatrix} )
= \begin{bmatrix} F & F & \mathbf{T} \\ F & \mathbf{T} & F \\ F & F & F  \end{bmatrix}    
\end{equation*}

With this definition of §score§ and §select\_best§ we obtain §open\_for\_close§ using the alternative approach described above (the last opener with the same depth as the closer's, that is still before the closer) as follows:

\begin{lstlisting}
possible_open_for_close = 
    select(indices,indices,<) and 
    select(opens,True,==) and 
    select(adjusted_depth,adjusted_depth,==);
open_for_close = select_best(
                        open_for_close,
                        score(indices,1) );
\end{lstlisting}
This approach does not use §depth\_index§ to obtain §open\_for\_close§, allowing us to save a layer in our calculation.

\subsection{Computation flows for select solutions}
{\rasp} can compile the the architecture of any \encoder, and display it with an example input sequence. The command is §draw(s2s,inp)§ where §s2s§ is the target \encoder~ and §inp§ is the example sequence to display, e.g., §draw(dyck1,"(())")§.

Example computation flows for §hist\_bos§ and §reverse§ are given in the main paper in Figures  \ref{fig:histbos} and \ref{fig:reverse}, respectively. 

\begin{figure*}
\centering
    \includegraphics[scale =0.26] {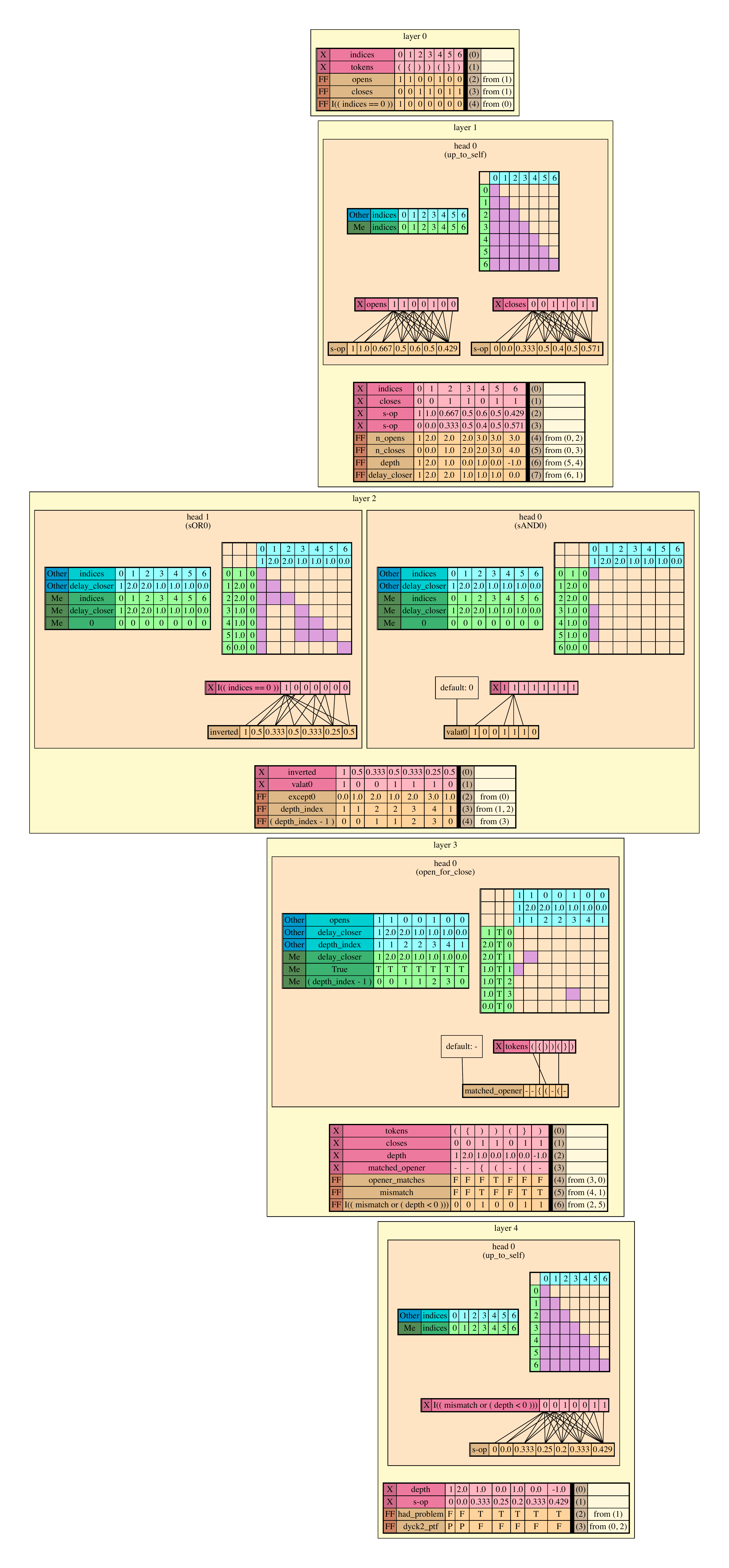}
\caption{Computation flow in compiled architecture from {\rasp} solution for Dyck-$2$, for solution shown in \Cref{app:rasp:dyck2}. Applied to the unbalanced and `incorrectly matched' (with respect to structure/pair-matches) sequence §"({))(})"§.
    }
    \label{fig:dyck2}
\end{figure*}

An example computation flow for §hist\_nobos§ is given in \Cref{fig:hist_nobos}. The double-histogram flow partially shown in \Cref{fig:accept} is shown in full in \Cref{fig:hist2_full}. Computation flows for the compiled architectures of §sort§ and for §most\_freq§ (as solved in Figures \ref{app:rasp:sort} and \ref{app:rasp:by_freq}) are shown in full, alongside the attention patterns of respectively attention-regularised transformers, in \Cref{Se:app:experiments}. Computation flows for Dyck-$1$-PTF and Dyck-$2$-PTF are shown in \Cref{fig:dyck1} and \Cref{fig:dyck2}.

\input{App2a_selector_width}

\begin{figure}
    \centering
    \begin{lstlisting}
reverse = aggregate(
    select(indices,
           length-indices-1,==)
    tokens );
    \end{lstlisting}
    \caption{{\rasp} one-liner for reversing the original input sequence, §tokens§. This compiles to an architecture with two layers: §length§ requires an attention head to compute, and §reverse§ applies a §select-aggregate§ pair that uses (among others) the \encoder~ §length§.}
    \label{app:rasp:reverse}
\end{figure}

\begin{figure}
    \begin{lstlisting}[mathescape=true,escapechar=\%]
def histf(seq, assume_bos = False) {
    same_tok = select(seq,seq,==);
    return selector_width(same_tok, 
            assume_bos= assume_bos);
}
    \end{lstlisting}
    \caption{{\rasp} program for computing histograms over any sequence, with or without a BOS token. Assuming a BOS token allows compilation to only one layer and one head, through the implementation of §selector\_width§ as in \Cref{fig:selector_width}.  The §hist\_bos§ and §hist\_nobos§ tasks in this work are obtained through §histf(tokens)§, with or without §assume\_bos§ set to §True§. }\label{app:rasp:hist}
\end{figure}

\begin{figure}
\begin{lstlisting}
def has_prev(seq) {
    prev_copy = 
        select(seq,seq,==) and
        select(indices,indices,<);
    return aggregate(prev_copy,1,0)>0;
}

is_repr = not has_prev(tokens);
same_count = 
    select(hist_bos, hist_bos,==);
same_count_reprs = same_count and
    select(isnt_repr, False,==);
hist2 =selector_width(
        same_count_reprs, 
        assume_bos = True);
\end{lstlisting}
\caption{{\rasp} code for hist-2, making use of the previously computed §hist§ \encoder~ created in \Cref{app:rasp:hist}. We assume there is a BOS token in the input, though we can remove that assumption by simply using §hist\_nobos§ and removing §assume\_bos=True§ from the call to §selector\_width§. The segment defines and uses a simple function §has\_prev§ to compute whether a token already has an copy earlier in the sequence. }\label{app:rasp:hist2}
\end{figure}

\begin{figure}
\begin{lstlisting}
def sort(vals,keys,assume_bos=False) {
    smaller = select(keys,keys,<) or 
        (select(keys,keys,==) and 
         select(indices,indices,<) );
    num_smaller = 
        selector_width(smaller,
            assume_bos=assume_bos);
    target_pos = num_smaller if 
                    not assume_bos else
    (0 if indices==0 else (num_smaller+1));
    sel_new = 
        select(target_pos,indices,==);
    sort = aggregate(sel_new,vals);
}
\end{lstlisting}
\caption{{\rasp} code for sorting the \encoder~ §vals§ according to the order of the tokens in the \encoder~ §keys§, with or without a BOS token. The idea is for every position to focus on all positions with keys smaller than its own (with input position as a tiebreaker), and then use §selector\_width§ to compute its target position from that. A further select-aggregate pair then moves each value in §val§ to its target position. The sorting task considered in this work's experiments is implemented simply as §sort\_input=sort(tokens,tokens)§.}\label{app:rasp:sort}
\end{figure}

\begin{figure}
    \centering
    \begin{lstlisting}[mathescape=true,escapechar=\%]
max_len = 20000;
freq = hist(tokens,assume_bos=True);
is_repr = not has_prev(tokens);
keys = freq - 
    indicator(not is_repr) * max_len;
values = tokens if is_repr else "$\bos$"
most_freq = sort(values,keys,
                 assume_bos=True);
    \end{lstlisting}
    \caption{{\rasp} code for returning the unique tokens of the input sequence (with a BOS token), sorted by order of descending frequency (with padding for the remainder of the output sequence). The code uses the functions §hist§ and §sort§ defined in Figures \ref{app:rasp:hist} and \ref{app:rasp:sort}, as well as the utility function §has\_prev§ defined in \Cref{app:rasp:hist2}. First, §hist§ computes the frequency of each input token. Then, each input token with an earlier copy of the same token (e.g., the second §"a"§ in §"baa"§) is marked as a duplicate. The key for each position is set as its token's frequency, minus the maximum expected input sequence length if it is marked as a duplicate. The value for each position is set to its token, unless that token is a duplicate in which case it is set to the non-token §\bos§. The values are then sorted by the keys, using §sort§ as presented in \Cref{app:rasp:sort}. }
    \label{app:rasp:by_freq}
\end{figure}

\begin{figure}
    \begin{lstlisting}[mathescape=true,escapechar=\%]
def num_prevs(bools) {
    prevs = select(indices,indices,<=);    
    return (indices+1) *
           aggregate(prevs,
                     indicator(bools));
}
n_opens = num_prevs(tokens=="(");
n_closes = num_prevs(tokens==")");
balance = n_opens - n_closes;
prev_imbalances = num_prevs(balance<0);
dyck1PTF = "F" if prev_imbalances > 0 
                    else
          ("T" if balance==0 else "P");
    \end{lstlisting}
    \caption{{\rasp} code for computing Dyck-$1$-PTF with the parentheses §(§ and §)§.}\label{app:rasp:dyck1}
\end{figure}

\begin{figure*}
\centering
    \includegraphics[scale =0.5,trim = 0 0 0 0] {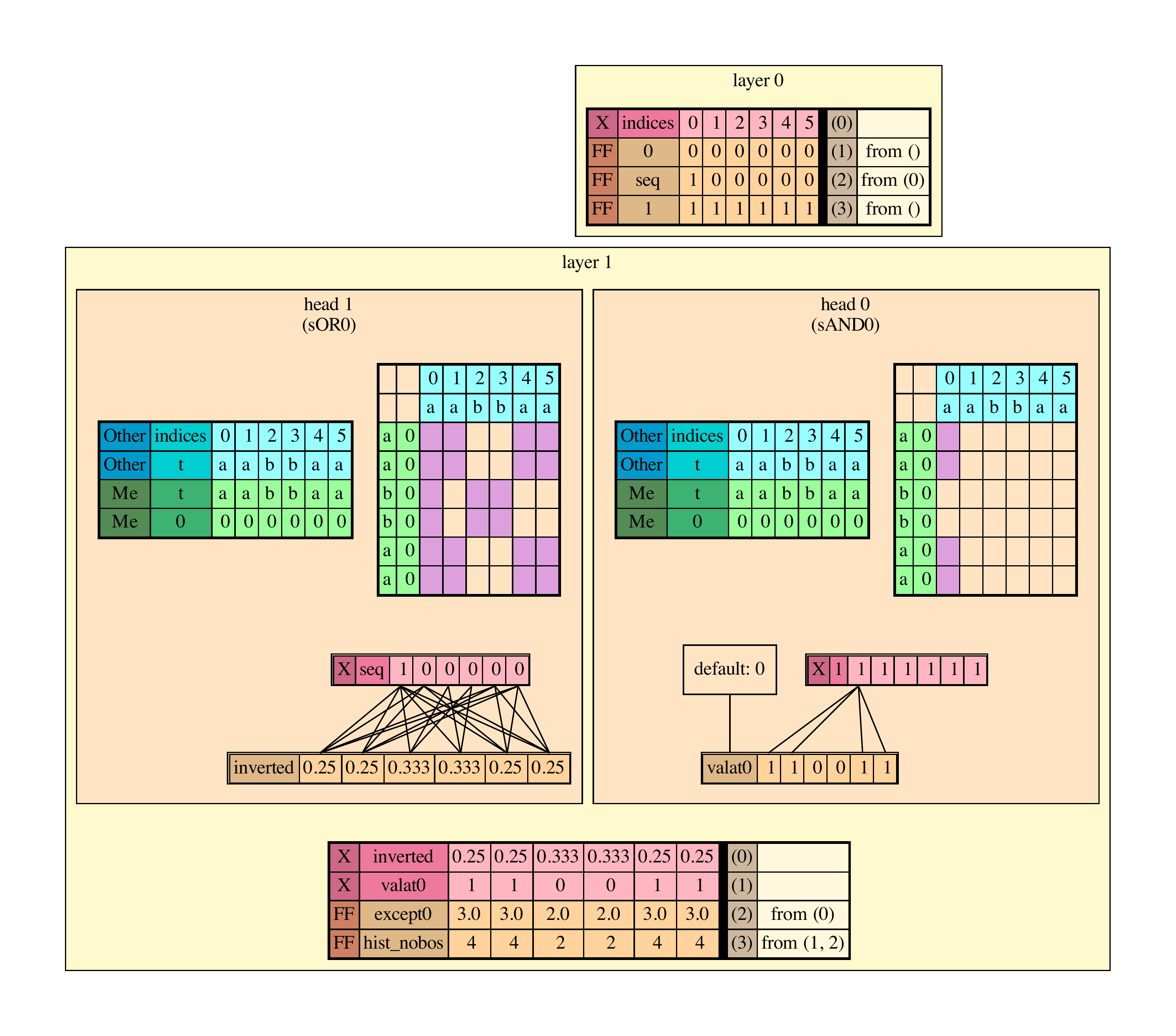}
\caption{Computation flow in compiled architecture from {\rasp} solution for histogram without a beginning-of-sequence token (using §histf(tokens)§ with §histf§ from \Cref{app:rasp:hist}). We present the short sequence §"aabbaa"§, in which the counts of §a§ and §b§ are different.}
    \label{fig:hist_nobos}
\end{figure*}

\begin{figure*}
\centering
    \includegraphics[scale =0.35,trim = 0 0 0 0] {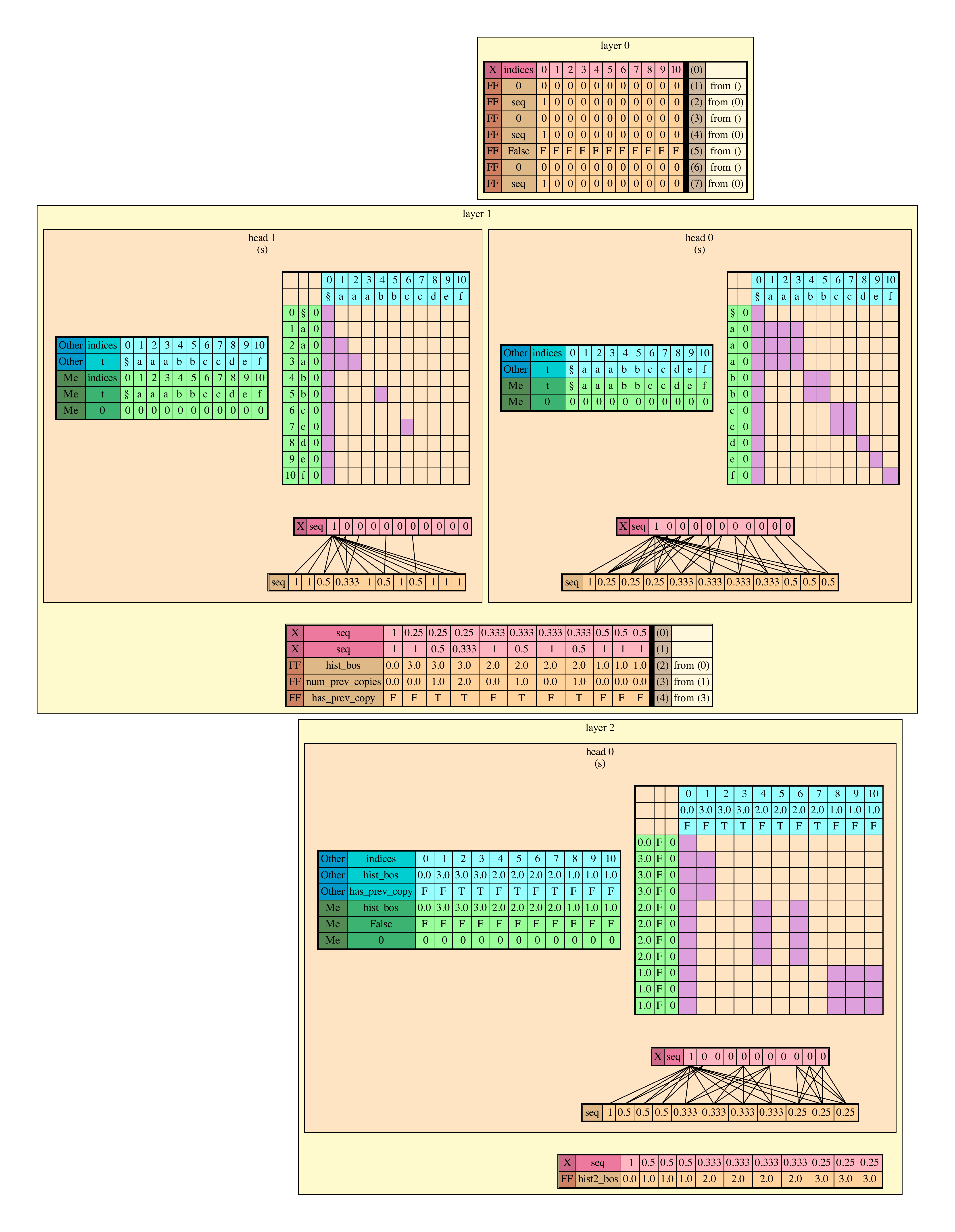}
\caption{Computation flow in compiled architecture from {\rasp} solution for double-histogram, for solution shown in \Cref{app:rasp:hist2}. Applied to §"\bos aaabbccdef"§, as in \Cref{fig:accept}.
    }
    \label{fig:hist2_full}
\end{figure*}

\begin{figure*}
\centering
    \includegraphics[scale =0.4] {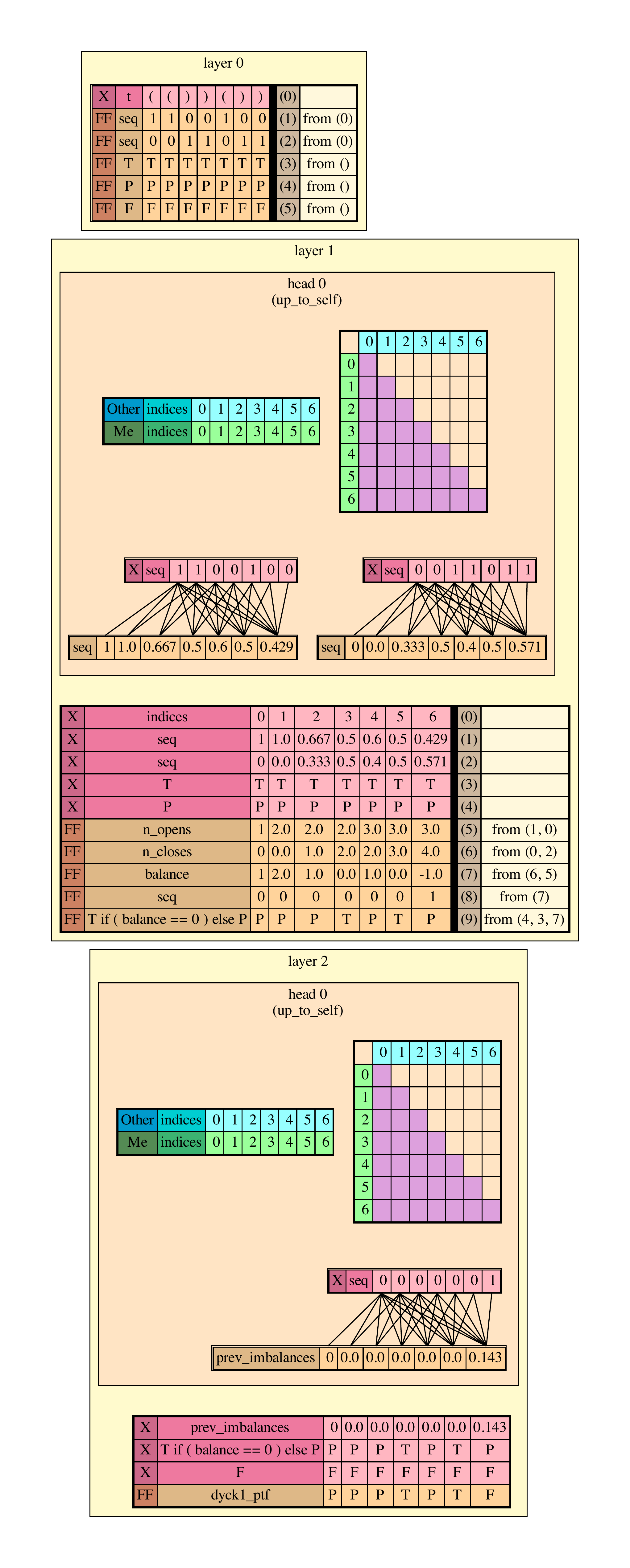}
\caption{Computation flow in compiled architecture from {\rasp} solution for Dyck-$1$, for solution shown in \Cref{app:rasp:dyck1}. Applied to the unbalanced input sequence §"(())())"§.
    }
    \label{fig:dyck1}
\end{figure*}

%% file: App2b_dyck2.tex
\begin{figure}
    \begin{lstlisting}[mathescape=true,escapechar=\%]
pairs = ["()","{}","[]"]; # etc ...
openers = [p[0] for p in pairs];
closers = [p[1] for p in pairs];
opens = tokens in openers;
closes = tokens in closers;
n_opens = num_prevs(opens);
n_closes = num_prevs(closes);

depth = n_opens - n_closes;
adjusted_depth = 
    depth + indicator(closes);
earlier_same_depth = 
    select(adjusted_depth,adjusted_depth,==) 
        and
    select(indices,indices,<=);
depth_index = 
    selector_width(earlier_same_depth);
open_for_close = 
        select(opens,True,==) and 
        earlier_same_depth and
        select(depth_index,
               depth_index-1,==);
matched_opener = 
    aggregate(open_for_close,tokens,"-");
opener_matches = 
    (matched_opener+tokens) in pairs;
mismatch = closes and not opener_matches;
had_problem = 
    num_prevs(mismatch or depth<0 )>0;
dyck3 = "F" if had_problem else 
	("T" if depth==0 else "P");
    \end{lstlisting}
    \caption{Pure {\rasp} code (as opposed to with an additional select-best operation) for computing Dyck-$3$-PTF with the parentheses §(,)§, §\{,\}§ and §[,]§. The code can be used for any Dyck-$n$ by extending the list §pairs§, without introducing additional layers or heads. }\label{app:rasp:dyck2}
\end{figure}

%% file: App2a_selector_width.tex
\begin{figure}
    \centering
        \begin{lstlisting}[mathescape=true,escapechar=\%]
    def selector_width(sel,
                assume_bos = False) {
                % %
        light0 = indicator(
                    indices == 0);
        or0 = sel or select_eq(indices,0);
        and0 =sel and select_eq(indices,0);
        or0_0_frac =aggregate(or0, light0);
        or0_width = 1/or0_0_frac;
        and0_width = 
            aggregate(and0,light0,0);
        
        # if has bos, remove bos from width
        # (doesn't count, even if chosen by
        #  sel) and return.
        bos_res = or0_width - 1;
        
        # else, remove 0-position from or0,
        # and re-add according to and0:
        nobos_res = bos_res + and0_width;
        
        return bos_res if assume_bos else 
                    nobos_res;
    }
        % %
\end{lstlisting}
    \caption{Implementation of the powerful {\rasp} operation §selector\_width§ in terms of other {\rasp} operations. It is through this implementation that {\rasp} compiles §selector\_width§ down to the transformer architecture.}
    \label{fig:selector_width}
\end{figure}